\documentclass[10pt,journal,compsoc]{IEEEtran}
\pdfoutput=1
\usepackage{xcolor}
\usepackage{verbatim}
\usepackage{color}
\usepackage{times}
\usepackage{epsfig}
\usepackage{graphicx}
\usepackage{amsmath}
\usepackage{amssymb}
\usepackage{xspace}
\usepackage{soul}
\usepackage{url}
\usepackage{tabularx}
\usepackage{adjustbox}
\usepackage{makecell}
\usepackage{caption}
\usepackage{siunitx}
\usepackage{graphicx}
\usepackage{enumitem}
\usepackage{mdframed}
\usepackage{booktabs}
\usepackage{xcolor, xspace}
\usepackage[ruled, lined, vlined, linesnumbered]{algorithm2e}
\usepackage{multirow, makecell}
\usepackage{threeparttable, tablefootnote}
\usepackage{tikz}

\newcommand{\ie}{\emph{i.e.}\xspace}

\newcommand{\eg}{\emph{e.g.}\xspace}
\newcommand{\etc}{\emph{etc.}\xspace}
\newcommand\figref[1]{Fig.~\ref{#1}}
\newcommand\tabref[1]{Table~\ref{#1}}
\newcommand\secref[1]{Section~\ref{#1}}
\newcommand\equref[1]{Eq.(\ref{#1})}
\newcommand{\fakeparagraph}[1]{\vspace{1mm}\noindent\textbf{#1.}}
\newcommand{\systemname}{SwapNet\xspace}
\newcommand{\sufficient}{DInf\xspace}
\newcommand{\compression}{TPrg\xspace}

\title{\systemname: Efficient Swapping for DNN Inference on Edge AI Devices Beyond the Memory Budget}

\begin{document}
\IEEEtitleabstractindextext{
\author{Kun~Wang, Jiani~Cao, Zimu Zhou,~\IEEEmembership{Member,~IEEE,} and~Zhenjiang~Li,~\IEEEmembership{Member,~IEEE} \\
}

\IEEEcompsocitemizethanks{\IEEEcompsocthanksitem K.~Wang, J.~Cao and Z.~Li are with the Department of Computer Science, City University of Hong Kong, Hong Kong, China. E-mail: kwang69-c@my.cityu.edu.hk, jncao2-c@my.cityu.edu.hk, zhenjiang.li@cityu.edu.hk.
\IEEEcompsocthanksitem Z.~Zhou is with the School of Data Science, City University of Hong Kong, Hong Kong, China. E-mail: zimuzhou@cityu.edu.hk.
}

\begin{abstract}
Executing deep neural networks (DNNs) on edge artificial intelligence (AI) devices enables various autonomous mobile computing applications. 
However, the memory budget of edge AI devices restricts the number and complexity of DNNs allowed in such applications.
Existing solutions, such as model compression or cloud offloading, reduce the memory footprint of DNN inference at the cost of decreased model accuracy or autonomy. 
To avoid these drawbacks, we divide DNN into blocks and swap them in and out in order, such that large DNNs can execute within a small memory budget. 
Nevertheless, naive swapping on edge AI devices induces significant delays due to the redundant memory operations in the DNN development ecosystem for edge AI devices.
To this end, we develop \systemname, an efficient DNN block swapping middleware for edge AI devices. 
We systematically eliminate the unnecessary memory operations during block swapping while retaining compatible with the deep learning frameworks, GPU backends, and hardware architectures of edge AI devices. 
We further showcase the utility of \systemname via a multi-DNN scheduling scheme.
Evaluations on eleven DNN inference tasks in three applications demonstrate that \systemname achieves almost the same latency as the case with sufficient memory even when DNNs demand $2.32\times$$\sim$ $5.81\times$ memory beyond the available budget. The design of SwapNet also provides novel and feasible insights for deploying large language models (LLMs) on edge AI devices in the future.
\end{abstract}
}
\maketitle

\section{Introduction}
\label{sec:intro}
\IEEEPARstart{T}{he} use of edge artificial intelligence (AI) devices, such as the NVIDIA Jetson series~\cite{mAI}, has rapidly grown in recent years for developing various autonomous applications~\cite{he2016deep,simonyan2014very,long2015fully, redmon2018yolov3}, such as self-driving cars, surveillance drones, vehicle-to-everything (V2X) infrastructure, \etc 
However, this progress comes with new challenges for \textit{memory management} when performing deep learning tasks on edge AI devices.
On the one hand, deep neural networks (DNNs) continuously grow in size and complexity, and many applications require multiple tasks to be performed concurrently~\cite{yang2019re,baidya2020vehicular}.
On the other hand, edge AI devices often have limited memory to accommodate all the tasks as well as for buffer cache and page cache to tolerate workload dynamics.
For example, a self-driving application contains a fleet of DNNs for lane detection, pedestrian recognition, scene segmentation and depth estimation.
It may also perform continuous SLAM navigation and frequent video capture \cite{durrant2006simultaneous}. 
Yet commercial off-the-shelf edge AI platforms, \eg, the RosMaster X3 autonomous vehicle~\cite{rosmaster}, are equipped with merely 2--8 GB memory for all these tasks as well as other system services, such as operating system and CUDA kernel.

Conventional solutions to efficient DNN inference with limited memory budget include model compression \cite{han2015deep,liu2018demand,park2022mgemm,xie2019source,hou2022neulens,polino2018model,elsken2019neural} and cloud offloading \cite{kang2017neurosurgeon,jiang2022primask,huang2022real,lee2019occlumency,li2019edge,wang2021context}. 
Model compression techniques remove redundant parameters or decrease the precision of model parameters in the DNN, but usually induce accuracy losses at high compression rates, which is undesired in mission-critical applications, \eg, self-driving.
Cloud offloading schemes execute part of or the entire DNN on the cloud, which is vulnerable to unpredictable latency and may raise data privacy concerns due to its reliance on network connections.

In this paper, we aim at \textit{memory-friendly} DNN execution on \textit{edge AI devices} from an \textit{orthogonal} perspective.
Inspired by the traditional virtual memory mechanisms, we propose to partition large DNN models into small blocks (each block consists of one or more neural network \textit{layers}), and \textit{swap} these blocks in and out of the limited memory in order for execution.
This swapping strategy would allow on-device execution of DNN models with sizes even beyond the memory budget on the edge AI device.

However, it is challenging to devise a block swapping mechanism for efficient DNN inference on edge AI devices due to their unique DNN development ecosystem (\ie, deep learning frameworks, GPU backends, and hardware architectures).
Specifically, the system supports for edge AI devices are optimized to unleash the potentials of GPUs without dedicated care for the memory footprint and memory architecture.
Consequently, naive implementation of block swapping leveraging standard APIs provided for edge AI devices incurs considerable delay and memory consumption.
There are memory-efficient deep learning frameworks (\eg, TFLite~\cite{tflite}, MACE~\cite{mace} and NCNN~\cite{ncnn}) and optimizations \cite{wang2021asymo, jia2022codl, jeong2022band, wang2022melon} designed for mobile devices.
Yet they are not directly applicable to edge AI devices because they do not support the high performance backend for edge AI devices, \eg, CUDA~\cite{cuda}.

To this end, we propose \systemname, a \textit{transparent middleware} that enables \textit{efficient DNN block swapping} while remaining \textit{compatible} with the mainstream DNN development ecosystem for \textit{edge AI devices}.
We identify a series of \textit{unnecessary memory operations} and \textit{redundant memory copies} as the efficiency bottlenecks when naively swapping blocks on top of the standard DNN development tool chain.
On this basis, we design a novel zero-copy swap-in scheme and a model assembly by reference strategy to bypass the unnecessary memory copying and processing during block swapping.
Furthermore, we demonstrate the usability of \systemname by seamlessly integrating it with upper-layer scheduling algorithms for efficient multi-DNN execution. We evaluate \systemname on commodity edge AI devices with a rich set of applications covering self-driving, road-side unit and UAV surveillance. 
Compared to the default model compression method \cite{tp} in PyTorch \cite{pytorch}, \systemname can achieve 2.4--7.3\% higher accuracy using even 35.7--65.7\% less memory. 
The latency increases by only 6.2\% on average compared to the execution of sufficient memory without swapping.

\begin{figure}[t]
    \centering
    \includegraphics[width=.45\textwidth]{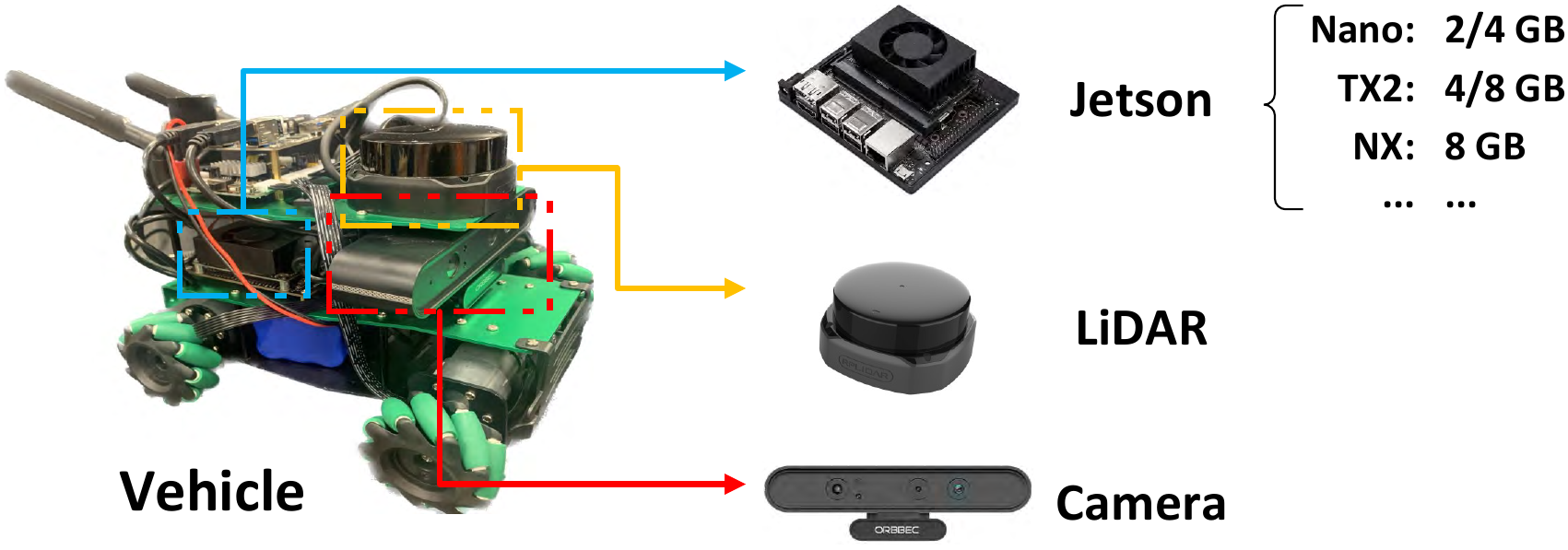}
    \caption{Illustration of an edge AI device based autonomous vehicle. It is expected to run multiple DNN and non-DNN tasks at a low memory budget.}
    \vspace{-.15in}
    \label{f:scenario}
\end{figure}

The significance of the \systemname design is to reveal a fundamental mismatch between the hardware architecture of edge AI devices and commercial deep learning frameworks. The inefficiencies in block swapping and model assembly described above are a specific manifestation of this problem. \systemname is designed to solve the root cause of the problem. Therefore, \systemname can provide new design insights for future development and research in the edge AI ecosystem. On the other hand, with the growing trend of deploying large language models (LLMs) or their smaller approximations (such as LLaMA-7B developed by Meta~\cite{llama}) in edge AI applications, the design of \systemname also provides novel and feasible insigts for deploying LLMs on edge AI devices in the future. In summary, we make the following contributions:
\begin{itemize}
    \item 
    We propose block swapping for efficient on-device execution of large DNNs within a small memory budget. 
    It is an orthogonal strategy to model compression and cloud offloading and is preferable for autonomous, mission-critical ubiquitous computing applications. Since the parameters and structure of the model itself are not changed, our design does not affect the model accuracy. In addition, it only relies on the device itself and does not affect the autonomy.
    \item 
    We develop \systemname, a new middleware for efficient block swapping on edge AI devices.
    It systematically eliminates the redundant memory copies and the associated operations with minimal modifications to the existing DNN development tool chain during DNN block swapping.
    To the best of our knowledge, \systemname is one of the first efforts in improving the memory efficiency of the unique DNN development ecosystem for edge AI devices.
    \item 
    We demonstrate the broad applicability of \systemname by combining it with a neat scheduling algorithm for efficient multi-DNN execution. Most advanced and effective scheduling algorithms can also integrate with \systemname according to the provided abstractions, to further optimize the real-time scheduling problem. 
    Extensive results on eleven DNN inference tasks in three application scenarios demonstrate promising performance gains compared to state-of-the-art alternative methods.
\end{itemize}

\vspace{-.2in}
\section{Preliminaries}
\label{sec:pre}
This section shows the challenges of DNN inference on edge AI devices.
We first demonstrate the limited memory budget to run typical DNN inference tasks on standard edge AI devices via a measurement study (\secref{sec:pre:memory}) and then highlight the uniqueness when designing memory-friendly DNN inference strategies on edge AI devices compared with mobile devices (\secref{sec:pre:edge}).

\subsection{Memory Budget of Edge AI devices for DNN Inference}
\label{sec:pre:memory}
There is a growing interest to deploy DNNs on edge AI devices for ubiquitous computing applications~\cite{baidya2020vehicular, barthelemy2019edge,xu2021limu}.
We illustrate the memory budget for DNN inference by developing a self-driving application on standard edge AI devices below.

\begin{itemize}
    \item \textbf{Setups.}
    We test the RosMaster X3 autonomous vehicle~\cite{rosmaster}, an off-the-shelf edge AI platform equipped with the NVIDIA Jetson device (see \figref{f:scenario}) to control the motors and on-board sensors such as LiDAR and depth cameras. 
    We also use the device in our evaluations (\secref{sec:eval}).
    The autonomous vehicle needs to run multiple \textit{DNN} and \textit{non-DNN} tasks to function properly.
    We measure the \textit{remaining memory} after executing the necessary \textit{non-DNN} tasks to understand the memory budget available for DNN inference.
    \item \textbf{Results.}
    \tabref{memoryusage} lists the memory usage of individual non-DNN tasks and the remaining memory for DNN tasks.
    Only $25\%$ out of the 8GB memory is left to simultaneously execute DNN tasks such as lane detection, pedestrian recognition, scene segmentation and depth estimation in autonomous driving scenario.
    The low memory budget severely constrains the number and complexity of DNN models to deploy on the autonomous vehicle platform.
\end{itemize}

\begin{table}[t]
\center
\small
\begin{tabular}{lll}
    \toprule
    Tasks & Memory Usage & Percentage  \\
    \midrule
    Operating System & 1038 MB & 12.7\%  \\
    SLAM and Navigation & 1815 MB & 22.2\% \\
    Map Repository & 1229 MB & 15.0\% \\
    Video Capture and Encoding & 488 MB & 5.9\% \\
    CUDA Kernel & 1518 MB & 18.5\% \\
    \midrule
    Remaining Memory & 2104 MB & 25.7\%  \\
    \bottomrule
\end{tabular}
\caption{Memory allocation of non-DNN tasks and the remaining memory budget for DNN tasks on an example autonomous vehicle application.}
\label{memoryusage}
\end{table}

\begin{figure*}[t]
    \centering
    \includegraphics[width=.9\textwidth]{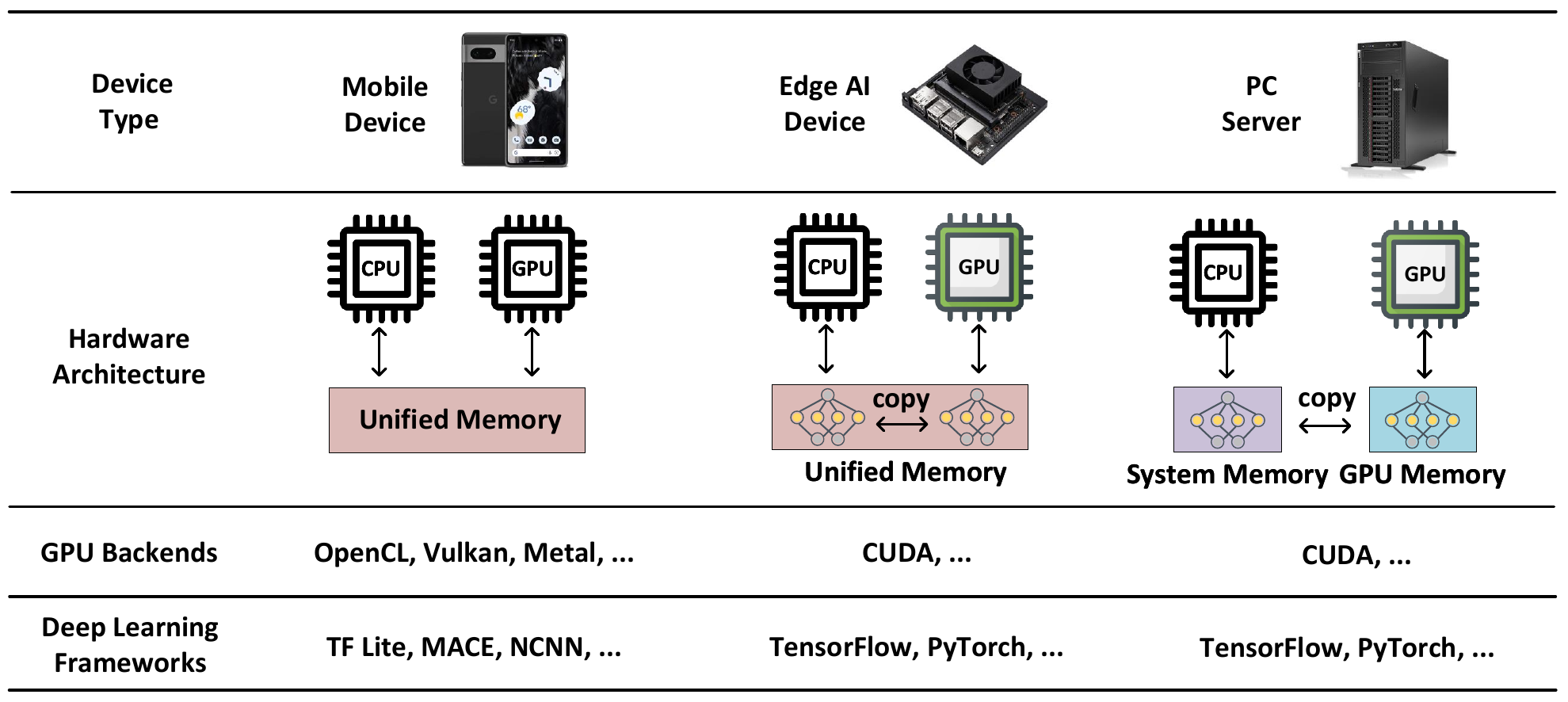}
    \caption{Comparison of the DNN development tool chain for mobile devices, edge AI devices, and the PC-grade devices.}
    \label{f:architecture}
\end{figure*}

\subsection{DNN Inference: Edge AI Devices vs. Mobile Devices}
\label{sec:pre:edge}

Despite extensive research on memory-efficient DNN inference~\cite{han2015deep,liu2018demand,li2019edge,wang2021context}, its applicability and the actual gains depend on the \textit{system support} of the targeting device categories \cite{sze2017efficient}. 
Here we explain the system support for DNN development on \textit{edge AI devices}, and highlight the differences from \textit{mobile devices}, another common device category in ubiquitous computing. The system support for DNN inference workloads roughly includes the \textit{deep learning framework}, \textit{GPU backend}, and the \textit{hardware architecture}. Figure~\ref{f:architecture} compares the corresponding system support for edge AI and mobile devices, as well as the desktop-version counterpart.
\begin{itemize}
    \item \textbf{Edge AI devices}.
    The typical DNN development ecosystem for off-the-shelf edge AI devices, \eg, the prevailing NVIDIA Jetson series, directly inherits the deep learning frameworks ported from the desktop versions, \eg, PyTorch and TensorFlow.
    This allows easy integration of the highly optimized backends, \eg, CUDA to unleash the full potentials of the underlying hardware, \eg, GPUs.
    However, the desktop-version deep learning frameworks are not optimized with stringent memory budget and overlook the unified memory architecture in mind, which induces significant overhead for block swapping due to redundant memory operations (see \secref{sec:swap:in} and \secref{sec:assemble:limitations}). For example, even if the CPU and GPU share the same memory in edge AI device, memory copy is still required when the GPU is called, which is redundant.
    \looseness=-1
    \item \textbf{Mobile Devices}.
    The DNN development chain for mobile devices, \eg, smartphones is specialized for the low resource platforms.
    For example, mobile deep learning frameworks such as TFLite~\cite{tflite}, MACE~\cite{mace} and NCNN~\cite{ncnn} often work with backends like OpenCL~\cite{opencl}, Metal~\cite{metal} and Vulkan~\cite{vulkan} to deploy DNNs to mobile GPU. Mobile devices tend to have only weak GPUs, designed for graphics rendering rather than complex matrix computing~\cite{jiang2020profiling}. Therefore, even though the mobile deep learning frameworks are optimized for memory-constrained platforms and unified memory architecture, they cannot be applied to edge AI devices because they do not support the high performance backend, \eg, CUDA backend.
    Directly applying these frameworks to edge AI devices would fail to make full use of the GPUs on edge AI devices.
\end{itemize}

\fakeparagraph{Summary}
There is a gap between \textit{exploiting the potential of GPUs} and \textit{optimizing the memory overhead} for DNN inference on edge AI devices.
On the one hand, the DNN development tool chain for edge AI devices optimizes for the full power of GPUs yet overlooks the memory budget and architecture.
On the other hand, applying memory-efficient deep learning frameworks designed for mobile devices to edge AI devices would under-utilize the capabilities of GPUs. 
This motivates us to devise a \textit{middleware} on top of existing DNN development ecosystem for edge AI devices for more efficient memory management while retaining full usage of the GPUs.

\section{\systemname Overview}
\label{sec:overview}

This section presents an overview of \systemname, a memory management middleware for efficient DNN inference on edge AI devices via swapping.
It enables lossless execution of multiple DNNs even beyond the memory budget on the edge AI devices.

\fakeparagraph{Design Scope}
\systemname is a \textit{transparent} memory management middleware for mainstream edge AI devices.
It aims to enable large DNN models inference beyond small memory budget. 
The design of \systemname accounts for the following features.
\begin{itemize}
    \item \textbf{Lossless}. 
    \systemname does not change the architecture and parameters of the DNN models, and thus avoids accuracy loss during DNN inference.
    \item \textbf{Transparent}.
    \systemname is seamlessly integrated with the mainstream deep leaning frameworks and users do not need to modify their code when using \systemname for memory-efficient DNN inference.
    \item \textbf{Lightweight}. 
    \systemname induces minimal modifications to the deep learning frameworks and operates with negligible memory overhead.
\end{itemize}
Note that we explain the operations of \systemname using the NVIDIA Jetson series~\cite{mAI}, the popular off-the-shelf edge AI computing platforms. 
They can make full use of powerful GPUs through the CUDA backend and the Pytorch deep learning framework. 
Since the DNN development tool chain for the NVIDIA Jetson series is almost identical to the desktop-grade devices, the NVIDIA Jetson series are widely used in various application scenarios.
However, \systemname can also be easily adapted to other edge AI devices such as Amazon DeepLens, Google Coral and Huawei Ascend.

\fakeparagraph{Basic Idea}
\systemname fulfills the above objectives via a \textit{swapping mechanism}. 
Swapping is a common memory management technique in modern operating systems to move data between memory and external storage when the system's physical memory is full. 
To enable large DNN inference beyond the memory budget, our idea is to partition the DNN model into \textit{blocks}, where each block consists of one or more neural network \textit{layers}. Given a memory budget $b$ allocated to a DNN model of size $s$ ($s>b$), we aim to partition it into $n$ blocks and execute them one by one. Therefore, we need to decide the number of blocks $n$ and the partition points, which we call the partitioning strategy. Once the partitioning strategy is obtained and the model is partitioned, the blocks are kept in the external storage, and are swapped in and out of the memory in order for DNN inference, which allows large DNNs to be executed within a small memory budget.

\fakeparagraph{Key Challenges}
Despite the simple idea, its key challenge is how to enable \textit{efficient DNN block swapping on edge AI devices in a transparent manner}.
A swapping mechanism consists of operations for block \textit{swap-in}, block \textit{assembly} for execution, and block \textit{swap-out}.
Naive block swap-in/out and assembly following the standard procedures in the DNN development tool chain for edge AI devices induces considerable delays and peak memory usage, due to \textit{redundant memory copies} and the associated operations.
It demands an in-depth analysis and dedicated designs to pinpoint and remove the inefficient bottlenecks for DNN block swapping with minimal modifications to the DNN development ecosystem.

\begin{figure}[t]
	\centering
	\includegraphics[width=.45\textwidth]{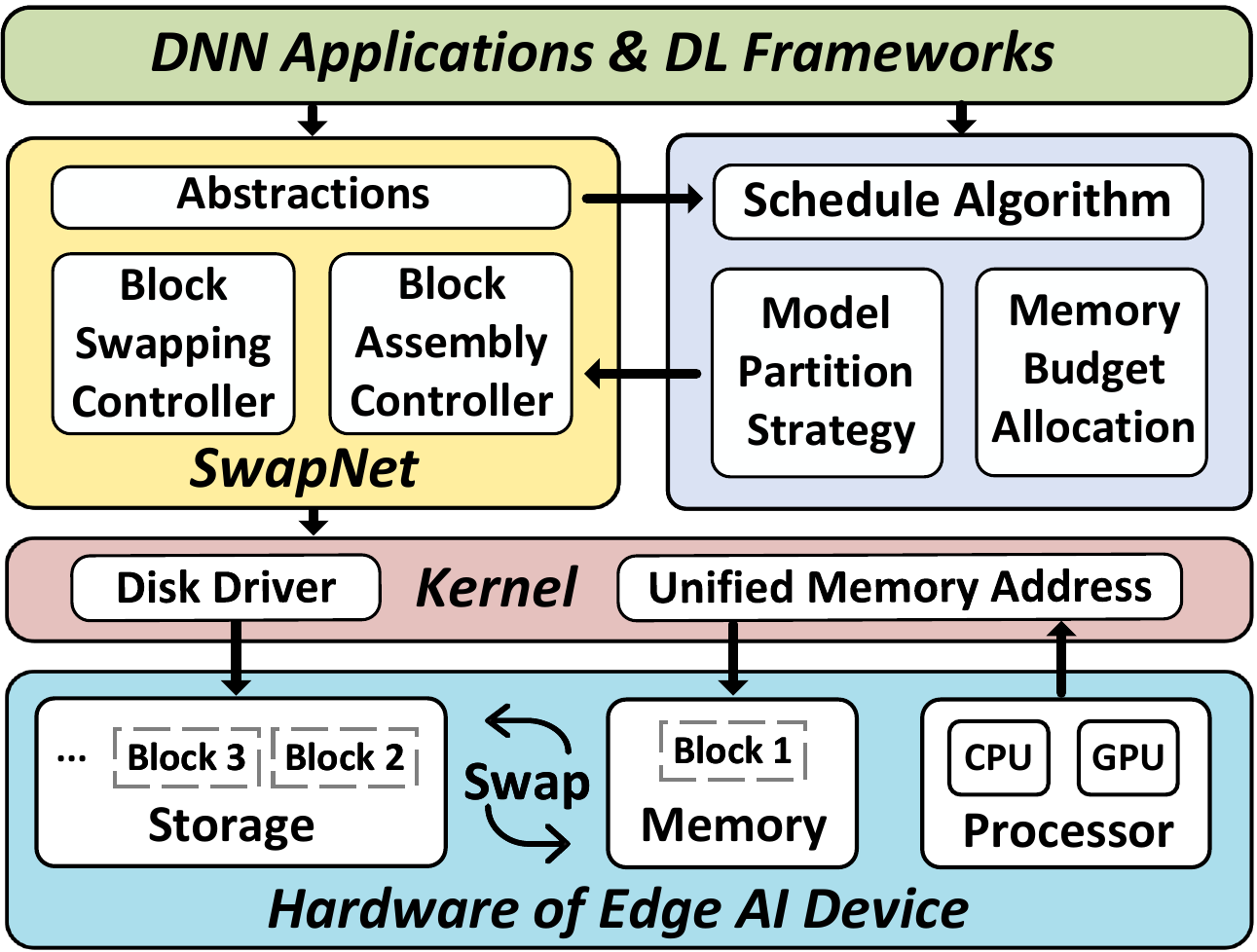}
	\caption{Overview of \systemname design.}
	\label{f:overview}
\end{figure}

\fakeparagraph{\systemname Architecture}
\systemname enables efficient DNN block (\ie, one or multiple layers) swapping on edge AI devices with two functional modules (see \figref{f:overview}). 
We assume the size of DNN model parameters exceeds the system memory budget and are initially kept in the external storage of the edge AI device.
\systemname allows efficient execution of the DNN model in blocks via the \textit{block swapping controller} and the \textit{block assembly controller}.
\begin{itemize}
    \item \textbf{Block Swapping Controller} (\secref{sec:swap}). 
    Since the primary swapping inefficiency lies in the unnecessary block copying into in-memory cache and fake GPU memory (\ie, actually the system memory) during swap-in (\secref{sec:swap:limitations}), we devise a block swapping controller that bypasses the copying via a novel zero-copy swap-in scheme (\secref{sec:swap:in}).
    \item \textbf{Block Assembly Controller} (\secref{sec:assemble}). 
    Since another inefficiency bottleneck comes from the use of a dummy model to assemble DNN parameters before execution (\secref{sec:assemble:limitations}), we replace the dummy model by pointers and apply an assembly by reference strategy to generate the executable objects (\secref{sec:assemble:reference}).
\end{itemize}

\fakeparagraph{\systemname Workflow}
Given a model block, \systemname efficiently swaps it into the memory, assembles the model parameters for execution, and frees the memory after execution.
As a transparent middleware, \systemname can be seamlessly integrated with various scheduling algorithms for the targeting applications.
We demonstrate the usage of \systemname via the case of \textit{multi-DNN execution} on edge CPU-GPU platforms, where the overall memory consumption exceeds the budget of the edge AI device (\secref{sec:partition}).

As shown in \figref{f:overview}, given multiple DNN inference tasks from the targeting application, \systemname provides the necessary abstractions for an advanced scheduling algorithm to determine how to allocate the limited memory budget to each DNN inference task, and generate partition strategy (including number of blocks and location of partitions).
According to the resulting scheduling plan, \systemname loads blocks into the system memory, dispatch them to the CPU or GPU efficiently without any redundant memory copy according to the unified memory address, and assemble the model parameters for execution.
\systemname enables fast and low-memory swapping and also supports parallel execution of multiple blocks from the same DNN inference task.
After execution, \systemname frees the memory by triggering standard garbage collection and is ready to take new blocks into the memory.

\section{\systemname Block Swapping Controller}
\label{sec:swap}
This section explains the block swapping controller of \systemname.
It has two functionalities.
\textit{(i) Swap-In}: load a model block from the external storage to the system memory on edge AI devices.
\textit{(ii) Swap-Out}: release the memory of the model block without writing block back to the external storage.
We show that the efficiency bottleneck of block swapping via standard I/O operations is the unnecessary memory copying during swap-in (\secref{sec:swap:limitations}), and our solution is a lightweight \textit{zero-copy block swap-in} scheme (\secref{sec:swap:in}). 
Finally, we illustrate how the block swapping controller works in action (\secref{sec:swap:summary}).

\subsection{Limitations of Standard Block Swapping}
\label{sec:swap:limitations}
We argue that the efficiency bottleneck of block swapping for DNN inference lies in its swap-in rather than swap-out.
This is because the model parameters do not change during DNN inference, and we can directly free the memory of each block after execution without writing it back to the external storage.
Such a write-back-free swap-out strategy only induces the latency for releasing memory, which is typically short (about $30$ ms in \secref{sec:partition}).
In contrast, swapping in a block using standard I/O is highly inefficient, as explained below.

In standard swap-in, to load the DNN block data from the external storage to the memory of CPU and GPU, the \texttt{read} operation is called to load block data into the system memory for CPU access. 
If the computation is assigned to the GPU, a \texttt{dispatch} function is further called to copy the block data from the CPU memory area to the GPU memory area.
Such a swap-in mechanism has two drawbacks.
\begin{itemize}
    \item 
    The \texttt{read} operation will copy the block to a page cache in the memory.
    Since multiple tasks share the limited memory, the page cache may experience high miss rate, and thus a long latency.
    Also, saving an extra copy of the block in memory contradicts to our goal to reduce the memory footprint.
    \item
    Due to the unified memory architecture, the GPU on edge AI devices do not have their dedicated memory (\ie, CPU and GPU physically shared the system memory yet logically separated), the \texttt{dispatch} function convert the block to GPU-compatible format and copy the same block to the fake GPU memory actually means there are two memory copies of block co-existing in the same physical system memory\cite{mittal2019survey}. This operation will introduce extra notable latency (see \secref{sec:eval}) and memory cost.
\end{itemize}
In short, the standard swap-in mechanism would keep \textit{two unnecessary copies} of the block in memory and induce \textit{non-negligible delay} due to CPU-to-GPU memory format conversion.

\begin{figure}[t]
	\centering
	\includegraphics[width=.45\textwidth]{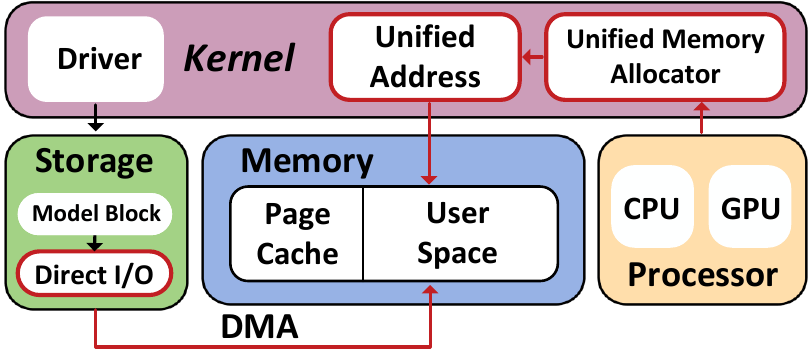}
	\caption{Workflow of block \systemname swap-in. Components in red are modifications on top of the standard block swap-in.}
	\label{f:dma}
\end{figure}

\subsection{Zero-Copy Block Swap-In}
\label{sec:swap:in}
To overcome the inefficiency of standard swap-in, we design a novel zero-copy block swap-in scheme.
It consists of a \textit{(i) direct block fetch} method to bypass the \texttt{read} operation and a \textit{(ii) copy-free GPU dispatch} function to eliminate the CPU-to-GPU memory conversion and copy overhead.

\subsubsection{Direct Block Fetch}
To avoid the problems of the page cache in the \texttt{read} operation, we leverage the direct memory access (DMA) and direct I/O to create a dedicated swap-in channel to fetch the blocks from the external storage to memory, as shown in \figref{f:dma}. 
Compared to using the page cache, the latency of this new swap-in channel is stable and it avoids an intermediate copy of the same block in the memory. 
The blocks swapped into memory via DMA are directly accessible by the CPU, or can be distributed to the GPU via the dispatch function, as we will discuss next.

\subsubsection{Copy-Free GPU Dispatch}
\begin{figure}[t]
	\centering
	\includegraphics[width=.5\textwidth]{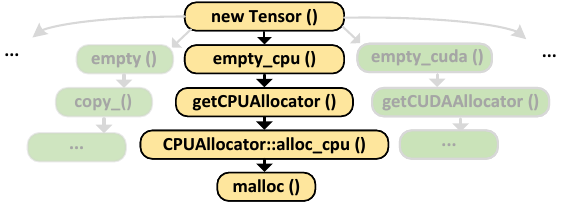}
	\caption{Dependence graph $G$ parsed in PyTorch to trace the location of CPU memory allocation.}
	\label{f:malloc}
 
\end{figure}

\begin{figure}[h]
    \centering	\includegraphics[width=.45\textwidth]{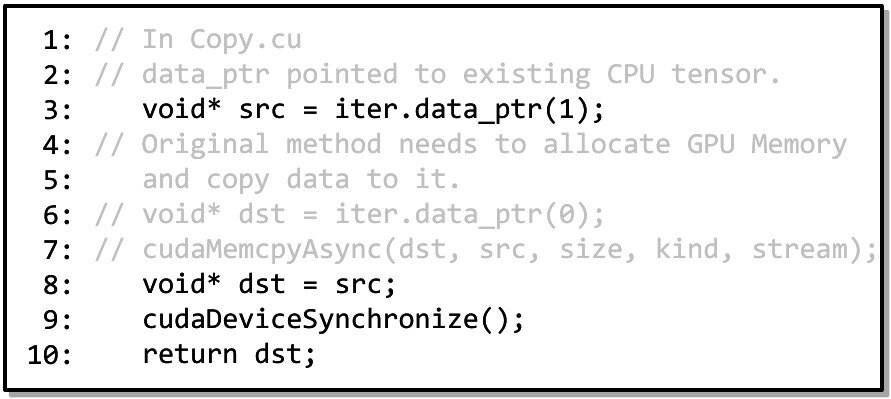}
    \caption{Code snippet for the revised GPU \texttt{dispatch} function.}
    \label{f:Code}
\end{figure}

Recall that the \texttt{dispatch} function, \eg, the \texttt{.to('cuda')} function~\cite{dispatch} needs to \textit{convert} the block to GPU-compatible format and \textit{copy} the block to the GPU memory for GPU using because the system memory is default allocated to the CPU but not GPU.
We propose to eliminate the format conversion and memory copying workload in GPU dispatch via \textit{allocating memory in unified addressing} and \textit{skipping the redundant memory copying} on top of existing deep learning frameworks for edge AI devices.
\begin{itemize}
    \item \textit{Allocate Memory in Unified Addressing}.
    The key reason for CPU-GPU memory format conversion is that the CPU and the GPU use separate logical memory addressing even though their memory is physically shared on edge AI devices.
    Accordingly, we can avoid memory format conversion if the memory allocation is conducted in the unified addressing.
    The deep learning frameworks for edge AI devices, \eg, PyTorch, allocate memory to CPU via the \texttt{malloc} function.
    Our idea is to replace it by the newly available \texttt{cudaMallocManaged} function, whose allocated memory can be accessed by both the CPU and the GPU \cite{UM}.
    To implement the idea, we trace the positions for CPU memory allocation in the deep learning framework, and replace \texttt{malloc} by \texttt{cudaMallocManaged} at the location with minimal modifications.
    For example, we can parse the source code of framework stack $\{src\}$ in PyTorch across the function calls related to keywords such as $\{\text{`}cpu\text{'}, \text{`}alloc\text{'}\}$, to derive the dependence graph $G$ of function calls for CPU memory allocation,
    \begin{eqnarray}
        parse(\{src\}, \{\text{`}cpu\text{'}, \text{`}alloc\text{'}\}) \rightarrow G \nonumber,
    \end{eqnarray}
    where the resulting dependence graph $G$ is shown in \figref{f:malloc}.
    Then we can replace the \texttt{malloc} at the bottom by \texttt{cudaMallocManaged}.
    The process applies to other deep learning frameworks such as TensorFlow~\cite{abadi2016tensorflow} and MNN~\cite{mnn}. 
    Afterwards, when blocks are swapped into memory, the memory is allocated in unified addressing, and there is no need for CPU-to-GPU memory format conversion. 
    \item \textit{Skip Redundant Memory Copying}.
    To avoid the memory copying while remaining compatible to the workflow of the deep learning framework stack, we revise the GPU \texttt{dispatch} function using the unified memory addressing.
    Originally, the \texttt{dispatch} function returns an address pointing to the newly allocated GPU-compatible space.
    Now, we modify it to return the block's address allocated in the unified address space after swapping in, and skip all other memory allocation and copy operations, as shown in \figref{f:Code}.
\end{itemize}

Since the new dispatch function avoids memory allocation and copy, the swap-in latency of blocks for GPU is almost the low as that for CPU (see \secref{sec:eval}).

\subsection{\systemname Block Swapping Controller in Operation}
\label{sec:swap:summary}
\figref{f:dma} illustrates the overall workflow of the block swapping controller in \systemname.
Blocks are swapped into the system memory for CPU execution (by default) using DMA and direct I/O. 
The allocated memory uses unified addressing so that it can be accessed by both the CPU and the GPU without memory format conversion.
If the DNN model is set for GPU execution, the revised GPU dispatch function is called. 
It only returns a pointer without memory allocation and copying.
Later after execution, the memory of the block is directly freed by garbage collection, and the released space is used for the next block to be swapped in.

\section{\systemname Block Assembly Controller}
\label{sec:assemble}
This section presents the block assembly controller of \systemname.
It assembles the model parameters (\ie, weights and bias) swapped into the memory according to the model architecture to be executable on the processors.
We show that frequent model assembly following the default procedure in deep learning frameworks is both time-consuming and memory-intensive, because it specifies the model architecture via a \textit{dummy model}, which takes up the same memory as the actual model (\secref{sec:assemble:limitations}).
Instead, we replace the dummy model by a \textit{skeleton}, which contains only pointers to the model parameters, and develop \textit{block assemble by reference}, an efficient block assembly strategy that works in synergy with frequent block swapping (\secref{sec:assemble:reference}).

\subsection{Limitations of Naive Block Assemble}
\label{sec:assemble:limitations}
\begin{figure}[t]
	\centering
    \includegraphics[width=.45\textwidth]{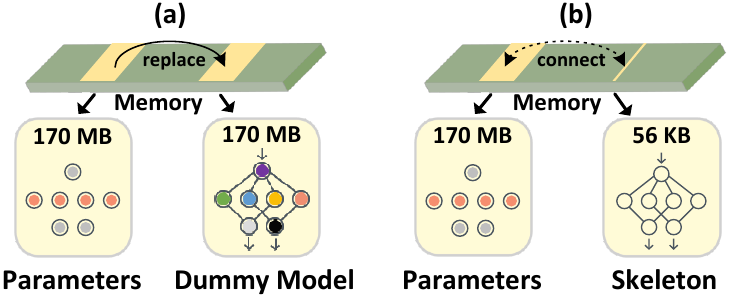}
	\caption{Illustration of (a) existing model assembly scheme and (b) our proposed assemble by reference strategy.}
	\label{f:restoration}
\end{figure}

By default, the network architecture information \ie, how tensors are connected, is stored during model object creation.
Specifically, a dummy model of the same network architecture yet with random parameters is generated as a memory placeholder.
To execute the model, its parameters are loaded into memory to replace the random weights in the dummy model, and the process is known as model assemble, as shown in \figref{f:restoration}(a).
In the context of swapping, the models are assembled and executed in blocks, yet the procedure remains the same.
There are two drawbacks of this default model assembly workflow in case of frequent block swapping.
\looseness=-1
\begin{itemize}
    \item 
    The dummy model blocks with random weights have the same size as the blocks of true model parameters. 
    This doubles the peak memory cost per block, which easily overwhelms the memory budget.
    \item 
    Model object instantiation and parameter-wise memory copy are necessary to replace the random weights in the dummy model, which incurs considerable delay.The operation is needed for every block swapping, leading to unacceptable latency. 

\end{itemize}

One naive solution is to configure the model in the \textit{inference} mode, because deep learning frameworks allow to serialize the parameters and the model configuration together and store them in a single file, which can be executed after de-serialization without model assembly. 
However, this solution has strong limitations. 
On the one hand, since the entire model has been compiled, it is easy to encounter errors that prevent the model from executing\footnote{For example, when the model is trained with several GPUs in cloud, then it cannot execute in the inference mode when it is deployed to an edge AI device with only one GPU due to configuration mismatch.}.
On the other hand, since the entire model is saved, the de-serialization process actually involves restoring the connection between the model skeleton and the weight, so de-serialization is also not fast, which is undesirable when blocks are frequently swapped.

\subsection{Block Assemble by Reference}
\label{sec:assemble:reference}
Since it is unnecessary to maintain a dummy model to specify the model architecture, we propose to only keep the model architectural information \ie, the skeleton, and directly refer to the true model parameters via pointers.
This would notably reduce the latency and memory footprint during model assembly.
Our model assembly by reference scheme works as follows.
\begin{itemize}
    \item \textit{Model Skeleton Extraction}. DNN models are described as objects in deep learning frameworks, \eg, \texttt{Obj\{pars, sket, attr\}}, where \texttt{pars} refers to the parameter of DNN models, \texttt{sket} describes the structure of the model, and \texttt{attr} means some unimportant attributes, such as the name and version of the model.
    We only keep the skeleton information in a model object, \ie, \texttt{Obj\{pars, sket, attr\}} $\rightarrow$ \texttt{Obj\{sket\}}.
    Note that \texttt{Obj\{sket\}} contains pointers only, which occupies no more than a few KB.
    Then we can serialize it for each model block and keep it in memory all the time. 
    The model parameters are still stored as a separate file \texttt{Fil\{pars\}} following the standard workflow. 
    Subsequently, after the corresponding \texttt{Fil\{pars\}} (\ie, a model block) is swapped into memory, \texttt{Obj\{sket\}} can be logically connected to \texttt{Fil\{pars\}}, as shown in \figref{f:restoration}(b). 
    \item \textit{Model Parameter Registration}. 
    To connect \texttt{Obj\{sket\}} and \texttt{Fil\{pars\}} with low latency, we store the model parameters in \texttt{Fil\{pars\}} as an array, and in \texttt{Obj\{sket\}} each pointer has the same index as its corresponding parameter in \texttt{Fil\{pars\}}.
    Therefore, we can simply iterate through the array, and write the address of each parameter in the corresponding pointer. 
    Since we do not need element search across the array, executable model blocks can be efficiently assembled through such address references.
\end{itemize}

\begin{figure*}[t]
    \centering
    \includegraphics[width=0.85\textwidth]{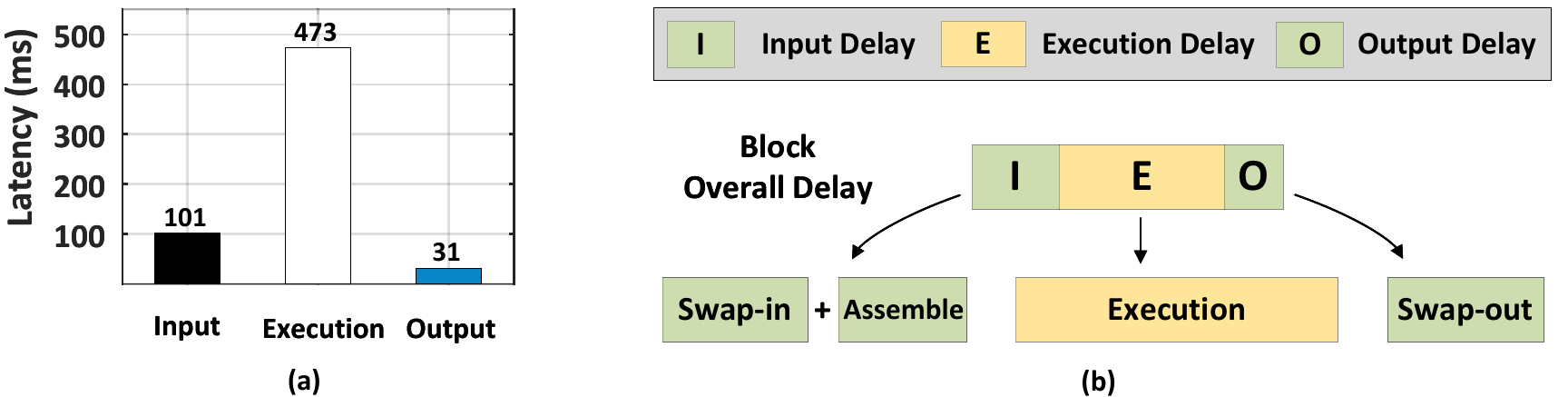}
    \caption{(a) The latency of three delay components in a ResNet-101 example. (b) What the three delay components contains.}
    \label{f:sequential}
\end{figure*}

\section{\systemname Utility: Multi-DNN Scheduling with Efficient Swapping} 
\label{sec:partition}
This section showcases the utility of \systemname as a transparent middleware.
Specifically, we explain how to harness the efficient swapping mechanism of \systemname for multi-DNN scheduling.
We first present the abstractions of \systemname provided to upper-level scheduling algorithms \secref{sec:app:abstraction}, and then illustrate how to devise a simple swapping-enabled multi-DNN scheduling scheme \secref{sec:app:scheduling}.

\subsection{\systemname Abstractions for Scheduling}
\label{sec:app:abstraction}
Given a model block $i$ with size $s_i$, parameter depth $d_i$, and \# of floating-point operations (FLOPs) $f_i$, \systemname provides \textit{three delay abstractions}, \ie, \textit{input delay $t_i^{in}$, execution delay $t_i^{ex}$, and output delay $t_i^{out}$} to the upper layer scheduling algorithms to determine how to partition and execute the given DNN models. \figref{f:sequential}(a) illustrates the latency of the three delay components in a ResNet-101 execution example. 
Note that the size $s_i$, parameter depth $d_i$, and FLOPs $f_i$ of a block are variables directly extracted from the given DNN architecture.
We explain how to derive the three delay components from these variables below combined with \figref{f:sequential}(b).

\begin{itemize}
    \item \textit{Input Delay $t_i^{in}$}.
    The input delay is the sum of the block swap-in delay $t^{in/sw}_i$ and block assembly delay $t^{in/as}_i$.
    The swap-in delay is proportional to the block size, \ie, $t^{in/sw}_i \propto s_i$, which is spent the input I/O for block $i$.
    The assembly delay is caused by address references.
    We empirically find that the delay of address references for different types, \eg, weights, biases and buffers, is consistent for the same edge AI device, which is around 50--55 $\mu$s by using a code profile tool~\cite{codeprofile}.
    Accordingly, the assembly delay is roughly proportional to the parameter depth of the block, 
    \ie, $t^{in/as}_i \propto d_i$.
    \item \textit{Execution Delay $t_i^{ex}$}.
    The execution delay of a block is proportional to its FLOPs, \ie, $t^{ex}_i \propto f_i$.
    \item \textit{Output Delay $t_i^{out}$}.
    The output delay is the latency due to block swap-out.
    In \systemname, the swap-out operations include resetting the pointers in the skeleton file to disconnect the model skeleton and parameters, and calling the garbage collector to release the memory occupied by the parameters. 
    The garbage collection delay can be considered as constant.
    Thus, the output delay is mainly depends on the time to reset the pointers, which is roughly proportional to the parameter depth, \ie, $t^{out}_i \propto d_i$.
\end{itemize}

\begin{figure*}[t]
    \centering
    \includegraphics[width=0.95\textwidth]{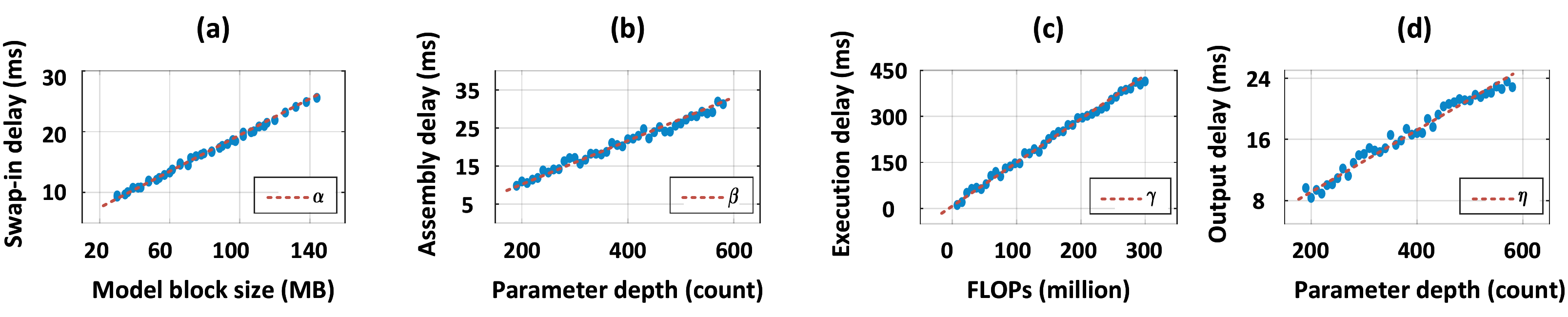}
    \caption{Profiling the four device-dependent coefficients ($\alpha$, $\beta$, $\gamma$, and $\eta$) for the three delay components via linear regression.}
    \label{f:fitting}
\end{figure*}

\begin{table}[t]
\center
\small
\begin{tabular}{llll}
    \toprule
    Layer    & Size     &  Depth & FLOPs \\
    \midrule
    Layer1     &   0.38 MB    &  1  &   26.2 M   \\
    Layer2     &   1.49 MB    &  5  &   0.8 K    \\
    Layer3     &   1.12 MB    &  1  &   123.9 M  \\
    Layer4     &   5.93MB     &  5  &   4.2 K    \\
    Layer5     &   4.38MB     &  6  &   316.7 M  \\
     ...       &     ...      & ... &   ...      \\
    Layer100   &   23.6 MB    &  1  &   30 K     \\
    Layer101   &   17.45 MB   &  1  &    5 K     \\
    \bottomrule
\end{tabular}
\caption{Example of ResNet-101 model information table.}
\label{modelinfotable}
\end{table}

In summary, the three delay components provided by \systemname can be estimated as $t^{in}_i = \alpha \times s_i + \beta \times d_i$, $t^{ex}_i = \gamma \times f_i$, and $t^{out}_i = \eta \times d_i$, where $\alpha$, $\beta$, $\gamma$, and $\eta$ are device-dependent coefficients, which can be easily profiled via linear regression (see \figref{f:fitting}).
Note that it is a one-off effort that can be conducted offline for the target edge AI device.
In \systemname, we profile a model info table for each DNN and store it as a meta file. 
This table is composed by the factors of each DNN layer. \tabref{modelinfotable} illustrate the example of ResNet-101 model information table.

\subsection{\systemname-Enabled Multi-DNN Scheduling Scheme}
\label{sec:app:scheduling}

Now we demonstrate how to design a multi-DNN scheduling scheme for edge AI devices upon the efficient swapping mechanism of \systemname.

\begin{figure}[t]
    \centering
    \includegraphics[width=.45\textwidth]{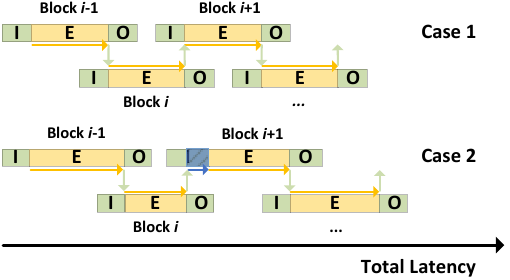}
    \caption{Two cases of parallel execution of blocks for a given DNN.}
    \label{f:parallel}
\end{figure}

\subsubsection{Efficient Multi-DNN Execution Problem} 
Given a set of DNN inference tasks, our objective is to minimize the maximum latency of these DNN inference tasks, where the total size of the DNNs may exceed the memory budget of the edge AI device.
The swapping mechanism of \systemname allows running these DNNs without accuracy loss beyond the limited memory budget.
As a transparent middleware, \systemname supports diverse upper-layer scheduling schemes.
For simplicity, we assume the following when executing the multiple DNNs.
\begin{itemize}
    \item 
    Each DNN is executed independently as an individual process to avoid interference across DNNs.
    \item
    Blocks of a single DNN can be executed in parallel to hide the latency of swap-in and swap-out.
\end{itemize}
A scheduling strategy that fulfills these requirements can be built upon \systemname to optimize the overall latency of these DNNs.
It should determine \textit{(i)} how to allocate the memory budget across multiple DNNs and \textit{(ii)} how to partition each DNN into blocks for efficient swapping and execution. 

To schedule the execution of multiple DNNs, we bind different DNN tasks into different CPU cores by setting the CPU affinity, and each DNN can execute independently as an individual process in specific CPU core. For the execution order of multiple DNNs, we execute all the DNN tasks concurrently. Due to the affinity of cores configured, multiple DNNs will not interfere each other. We now introduce a scheme below to instrument how to perform DNN scheduling on top of \systemname.

\subsubsection{Scheduling Scheme Design}  
\label{sec:app:memoryallocation}

Now we present a practical strategy for the multi-DNN execution problem above and explain how to configure the scheduling scheme according to the abstractions from \systemname.

\fakeparagraph{Allocate Memory Budget across DNNs}
If the total memory required by all the DNNs is within the memory budget, we directly allocate the memory required for each DNN.
If the total memory required exceeds the memory budget, we allocate the memory budget for the given DNNs as follows.

\textbf{\textit{1) Principle}}.
When we allocate the  memory budget, in addition to the inference latency and memory usage of each DNN (\secref{sec:app:abstraction}), we also consider the model complexity, which refers to whether a model can be easily divided into blocks. This is because our method needs the partition locations to schedule the blocks and optimize the latency. More partition locations are likely to bring more optimization opportunities. Earlier models such as such as VGG due to its simple structure are easy to partition and running fast, yet consume large memory, while more recent models like ResNet are more memory-efficient but are more difficult to partition and have slower inference speed (due to residual connections). Therefore, we introduce a performance score $PS=u*\frac{latency}{memory}$ for the memory budget allocation, where $u$ is a urgency degree of the DNN task. An urgent task will have larger $u$ according to the user configuration.
DNN tasks with a high performance score typically represent a complex structure, often characterized by smaller memory usage but longer inference latency. In contrast, those with a low performance score generally adopt a simpler structure, incurring larger memory usage but facilitating smaller inference latency.
Therefore, the DNN with higher $PS$ tend to obtain more memory budget and vice versa. 

\textbf{\textit{2) Strategy}.}
Consider $n$ models with the corresponding memory demand $\sum_{i}^{n}M_i$ and the available memory $M$, where $M$ $<$ $\sum_{i}^{n}M_i$. The actual memory $A_i$ allocated to model $i$ is:
\begin{equation}
    \small\label{eq:memory}
    A_{i}=\left( \frac{M_i}{\sum_{i=1}^n{M_i}}\right) \times \left(1 - \frac{1}{n}\right) \times M + \left( \frac{PS_i}{\sum_{i=1}^n{PS_i}}\right) \times \frac{1}{n} \times M,
\end{equation}
The rationale of this memory allocation strategy is two-fold.
\begin{itemize}
    \item 
    The memory is mainly allocated proportional to the ratio of the required memory (first term in Equation~\ref{eq:memory}). For example, if the $n$ models require 1.5GB memory (i.e., $\sum_{i}^{n}M_i=1.5\text{GB}$) while the available memory $M=1\text{GB}$, then model $i$ could be allocated with $\frac{M_i}{\sum_{i=1}^n{M_i}}\times M = \frac{2}{3}*{M_i} $ memory, which means with the memory constraint, each model is only assigned with 66.7\% memory budget. However, we only apply this strategy for $\left(1 -  \frac{1}{n}\right)$ of the available memory, where we reserve $\frac{1}{n}$ available memory as below.
    \item
    We reserve $\frac{1}{n}$ available memory and use it to calibrate the allocation according to the performance score (second term in Equation\mbox{~\ref{eq:memory}}). As mentioned earlier, a high performance score indicates a more complex model, and thus needs slightly more memory budget. We thus use the reserved $\frac{1}{n}$ of available memory for such refinement, and allocate memory to model $i$ according to its normalized performance score $\frac{PS_i}{\sum_{i=1}^n{PS_i}}$.
\end{itemize} 

\begin{table}[t]
\center
\small
\begin{tabular}{ccc}
    \toprule
    Partition Points     & Maximum Memory     &  Predicted Latency \\
    \midrule
    1,2     & exceed   & null      \\
    1,3     & exceed   & null      \\
    ...     & ...      & ...      \\
    30,66   & 105~MB    & 496~ms     \\
    30,67   & 109~MB    & 488~ms     \\
    ...     & ...      & ...       \\
    98,100  & exceed   & null      \\
    99,100  & exceed   & null      \\
    \bottomrule
\end{tabular}
\caption{Example of 3-blocks ResNet-101 run-time lookup table.}
\label{lookuptable}
\end{table}

\fakeparagraph{Partition Blocks within DNN}
Given a memory budget $b$ allocated to a DNN model of size $s$, we aim to partition and execute it in $n$ blocks to minimize the latency of the $n$ blocks.
The key optimization is to allow parallel execution of $m$ blocks to hide the latency of swap-in and swap-out.
\figref{f:parallel} illustrate the case for two blocks in parallel.
As next, we introduce how to determine the number of blocks $n$ to partition and how to partition the DNN into $n$ blocks in our design.

\begin{itemize}
    \item \textit{Determine number of blocks $n$ to partition}.
    If we allow $m$ blocks to be executed in parallel, then the total memory footprint of the blocks in execution should be within the memory budget, \ie, $m \times \frac{s}{n} \leq b$, which gives us $n = \left \lceil \frac{m \times s}{b}  \right \rceil$.
    We set $m=2$ in our scheduling scheme, since a higher order of parallelism, \ie, $m>2$ often leads more thread switching overhead. 
    \item \textit{Determine a partition scheme $p$}.
    A partition scheme $p = \{p_1,p_2,\ldots,p_{n-1}\}$ divides a given DNN into $n$ blocks. 
    As mentioned in  \secref{sec:overview}, a block consists of one or multiple layers.
    Hence, $p_{i}$ represents the layer indices of the DNN.
    Given a parallelism of $m=2$, the partition scheme $p$ aims to minimize the latency $t(p)$ to execute the $n$ blocks of the DNN.
    \begin{eqnarray}
         &   \arg\min\limits_{p} t(p),  \label{eqn:totaltime} \\
    \textit{s.t.} &  s_{i} + s_{i+1} \leq b \times (1 - \delta), ~\forall i \in [1, n-1], \label{eqn:constraint}
    \end{eqnarray}
    where $s_{i}$ is the size of block $i$.
    The $\delta$ refers to the reserved memory, contains the model skeleton, inference activation and look up table of each model.
    We can reformulate the objective in \equref{eqn:totaltime} into 
    \begin{eqnarray}
    \label{eqn:tp2}
        & \arg\min\limits_{p} \sum\nolimits_{i=2}^{n} \max\{t^{ov}_i(p),\,0\},
    \end{eqnarray}
    where $t^{ov}_i(p) = (t^{out}_{i-1}(p) + t^{in}_{i+1}(p)) - (t^{ex}_i(p) + t^{ov}_{i-1}(p))$, which is the residual delay that cannot be covered by the execution delay of block $i$.
    Note that \systemname has already profiled the delays for input, execution, and output, \ie, $\{t^{in}_{i}\}$, $\{t^{ex}_{i}\}$, and $\{t^{out}_{i}\}$ for any block size $s$ offline via linear regression (see \secref{sec:app:abstraction}).
    One can then apply any algorithms to derive a partition scheme $p$ that optimizes \equref{eqn:tp2}.
    We adopt a lookup table to derive the partition scheme $p$ in our implementation. 
    Specifically, we use the model info table and coefficients provided by \systemname to calculate the predictive delay of all candidate partition strategies for each model in the preparation stage and stored them in a lookup table, as shown in \tabref{lookuptable}. 
    At run-time, the search space is first pruned according to the allocated memory budget, and then we choose the strategy with least delay as the optimal partition strategy.
    These model blocks are constructed in a way that allows them to be executed sequentially or in parallel, with the output of one block feeding into the next, facilitating a modular and adaptable execution flow.
    Through this approach, the neural network is adaptively segmented into manageable blocks, each handled independently, which offers a flexible and adaptive execution strategy.
\end{itemize}

\fakeparagraph{Adaptively Partition and Exchange Blocks}   
In SwapNet, we also implement adaptive partitioning and exchange blocks of different DNNs. To adjust the model blocks efficiently, we do not divide the DNN model from scratch every time, which is very slow. Our strategy is that when the DNN model is first registered in the system, we first extract each layer of the DNN model, which can be regarded as the smallest block that can be divided from the DNN model. We then leverage these layers to form corresponding model blocks based on memory and latency requirements. In this process, we only need to adjust the index of the layer that each block needs to reference, so the adaptation of model blocks can be completed very quickly. Therefore, we design the following operations in \systemname:
\begin{itemize}
    \item \textbf{1) Initial layer-wise model division.} We design a function \texttt{get\_layers(Net)}, which separates all layers from the DNN model and returns the a sequence of layers extracted from the DNN model, which is a one-time effort for each DNN model.

    \item \textbf{2) Determining block-wise partition points.} The upper-layer scheduling algorithm then determines how to assemble blocks from each layer. The goal of this step is to ensure that the formed blocks can run independently while meeting resource and performance requirements.

    \item \textbf{3) Generating model blocks.} With the identified partition points, the function we designed \texttt{create\_blocks (part\_points, name, Layers)} is used to create the corresponding model blocks, where \texttt{part\_points} represents the partition points, telling SwapNet how to assemble each block. Each block is intended to be an independent entity, capable of performing inference independently once loaded with their weights.
\end{itemize}
When adaptation is required due to changes in the resource budget, we can simply repeat operations 2) and 3) above to generate new blocks, which can be completed in around 60-70 ms.

\section{Implementation}
\label{sec:impl}

\fakeparagraph{Implementation of \systemname} 
We implement \systemname (\secref{sec:swap} and \secref{sec:assemble}) using Python 3.6, C++ 14, CUDA 10.2 \cite{cuda} and PyTorch 1.6.0 (C++ version)~\cite{pytorch}.
We use the Linux kernel's DMAEngine~\cite{dma} to implement the DMA driver and employ the ``O\_DIRECT'' flag in the \texttt{open()} function to enable Direct I/O. 
To realize the memory allocation with unified addressing, we use \texttt{cudaMallocManaged} to replace the original CPU memory allocation, and link CUDA to the \texttt{CPUAllocator.cpp} source code in the file of \texttt{CMakeList} by using NVCC \cite{nvcc} to compile. 
We use the NVIDIA Visual Profiler \cite{nvvp} to check whether memory copying occurs for GPU inference.

\fakeparagraph{Implementation of Multi-DNN Scheduling Scheme} 
For the multi-DNN scheduling scheme (\secref{sec:partition}), we store the skeleton information of each model block and the table of feasible partition strategies as meta files. 
These files are kept in the memory during model execution and accounted for the budget overhead $\delta$. 
As for the execution order of multiple DNNs, we execute all the DNN tasks concurrently by using C++ multiprocessing. Parallel execution of model blocks is achieved through multithreading, and CPU affinity is fixed. Due to the affinity of cores configured, multiple DNNs will not interfere each other under our implementation.

\fakeparagraph{Discussions on Transparency} 
\systemname is implemented as a user-transparent middleware.
After installing it as a library, the relevant functions will automatically switch to our modified version when calling the GPU kernel. 
Users do not need to modify any of their codes. 
Specifically, we use \texttt{ctypes} to encapsulate the modified source code into the dynamic link library (\eg, \texttt{.so} in Linux and \texttt{.dll} in Windows) and dynamically load and link it at run time. 
This allows easy adoption of \systemname without recompiling the deep learning framework. 

\section{Evaluation}
\label{sec:eval}

\subsection{Experimental Setups}
\label{sec:exp:setups}
\subsubsection{Application Scenarios}
We test three real-world application scenarios involving 11 DNN inference tasks. 
\begin{itemize}
    \item \textbf{Self-Driving}.
    This scenario consists of the following tasks.
    1) YOLO v3 (236 MB) for object detection, 2) FCN (207 MB) for scene segmentation, 3) VGG 19 (548 MB) for traffic sign classification, and 4) ResNet-101 (170 MB) for forward car recognition. 
    The number after each model is the model size. 
    In addition to these DNN inference tasks, the scenarios also involves non-DNN tasks, such as operating system (OS) kernel, CUDA kernel, SLAM and navigation, video capture and transmission.
    \item \textbf{Road-Side Unit (RSU)}.
    This scenario includes the following tasks.
    1) two YOLO v3 (236 MB $\times 2$) for object detection on two on-board cameras, 2) two ResNet-101 (170 MB $\times 2$) for natural scenes classification, and 3) one VGG 19 (548 MB) for traffic light classification. 
    This application captures cases to execute a replica of a DNN model to process data from multiple sensors. 
    Similarly, the scenario also involves non-DNN tasks, such as multi-stream video capture and networking.
    \item \textbf{UAV Surveillance}.
    This scenario involves the following tasks.
    1) fire source detection with YOLO v3 (236 MB), and 2) wild animal recognition with ResNet-101 (170 MB). 
    Non-DNN tasks include OS kernel and HD video capture and transmission.
    In this scenario, we consider relatively ample resource budgets and investigate the utility of \systemname. 
\end{itemize}

\subsubsection{DNN Model Setup and Deployment} 
For each application scenario, we measure the average memory consumption of each non-DNN task, based on which we determine the total memory available to DNN tasks. We train VGG using GTSRB, and train ResNet with CIFAR100. We also train YOLO and FCN using COCO. 
We launch our scheduling scheme (\secref{sec:app:scheduling}) to determine the memory budgets and partition strategy accordingly. 
We just want to show the benefits from the combination of \systemname and proper scheduling algorithms. 
In practice, it can also be replaced by any more advanced and effective scheduling algorithm.
After the model is partitioned, the parameters of each block are stored in a SAMSUNG 970 evo plus NVMe SSD installed on the given edge AI device. 
In the evaluation, we configure VGG and ResNet to execute on CPU, and YOLO and FCN to execute on GPU based on the complexity of tasks.
\looseness=-1

\subsubsection{Edge AI Devices}
We mainly experiment with Nvidia Jetson NX with 8 GB memory, 1.9 GHz CPU (Nvidia Carmel) and 1.1 GHz GPU (Nvidia Volta). 
We also deploy and test \systemname on a Jetson Nano with a lower-end configuration, consisting of 4 GB memory, 1.4 GHz CPU and 0.6 GHz GPU, and further conduct a case study using a RosMaster X3 autonomous vehicle, equipped with Jetson NX and LiDAR and depth camera sensors.

\begin{figure}[t]
	\centering
	\includegraphics[width=3in]{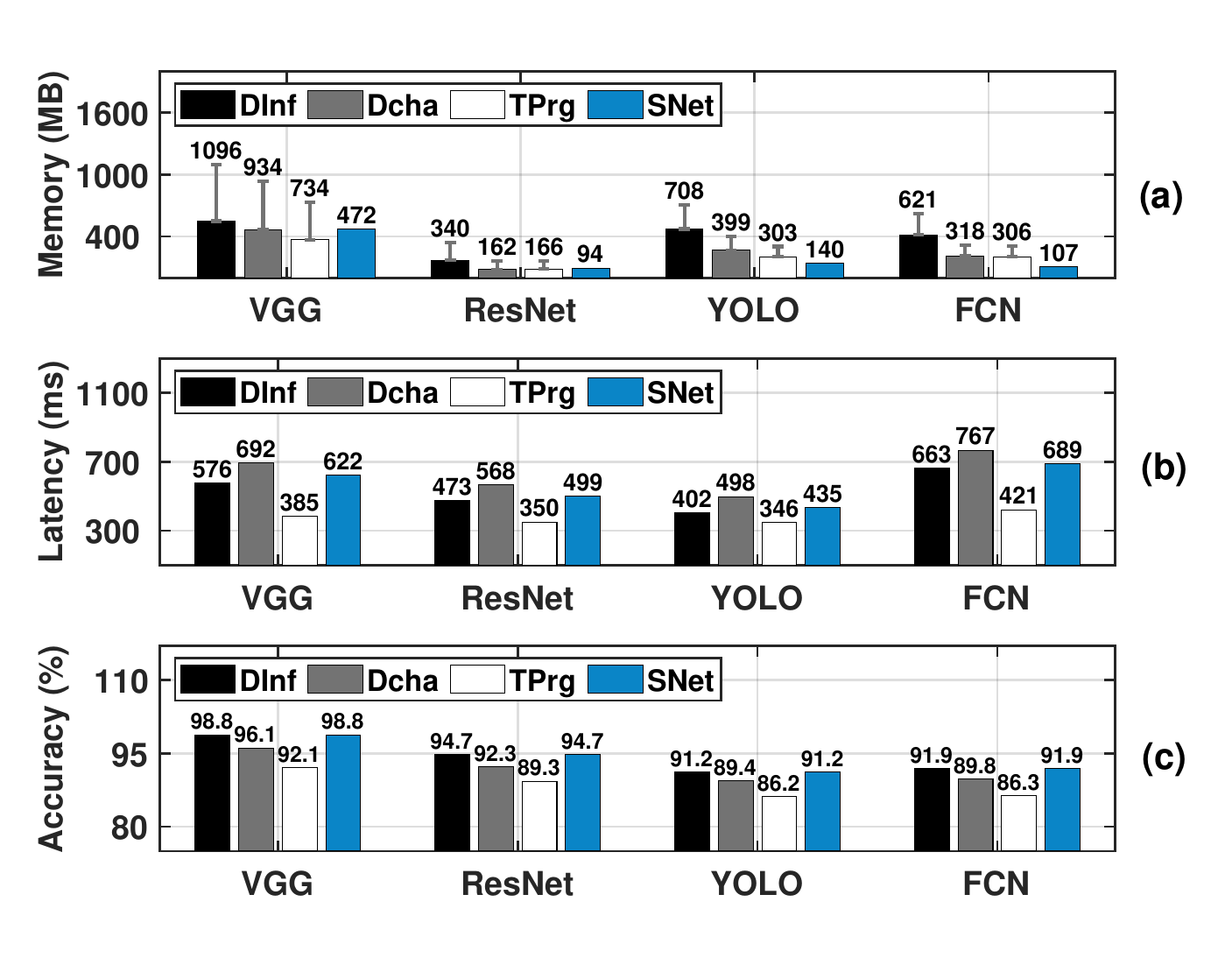}
	\caption{(a) Memory, (b) latency and (c) accuracy of each model in the self-driving application. Gray line in (a) indicates peak memory consumption.}	
	\label{f:overall1}
\end{figure}

\vspace{-.1in}
\subsection{Overall Performance}
We compare the performance of the following methods.
\begin{itemize}
    \item \textbf{Direct Inference (DInf)}.
    It executes DNN models directly without partitioning and is configured to terminate other non-DNN tasks when memory becomes tight. This is an ideal method to provide the best latency performance possible without loss of accuracy.
    \item 
    \textbf{Dividing By Channel (DCha)}.
    It is a type of state-of-the-art methods~\cite{hou2022dfsnet} that divide channels into groups to reduce memory consumption. For a fair comparison, all the divided channels are executed on the same device. 
    \item \textbf{Torch-Pruning (TPrg)}.
    It is a state-of-the-art model compression method~\cite{tp} that reduces model size to fit memory budget at the expense of accuracy loss.
    \item \textbf{\systemname (SNet)}.
    It is our proposed method.
\end{itemize}

We present the performance of these methods for each application scenario below.

\fakeparagraph{Performance on Self-Driving}
The memory consumption in this application is the same as the example in \secref{sec:pre}, where 843 MB space (not including 32 MB allocated as reserved memory $\delta$) is allocated to accommodate four DNN models, which is 1161 MB in total. According to our simple memory budget allocation algorithm in \secref{sec:app:memoryallocation}, the budgets for VGG, ResNet, YOLO and FCN are set to 475, 102, 142 and 124 MB, respectively\footnote{The VGG structure is highly unbalanced, largest layer takes up 392 MB, and a relatively large budget is required.}.
\compression compresses them to 367, 83, 101 and 102 MB and directly executes them.

\begin{figure}[t]
	\centering
	\includegraphics[width=3in]{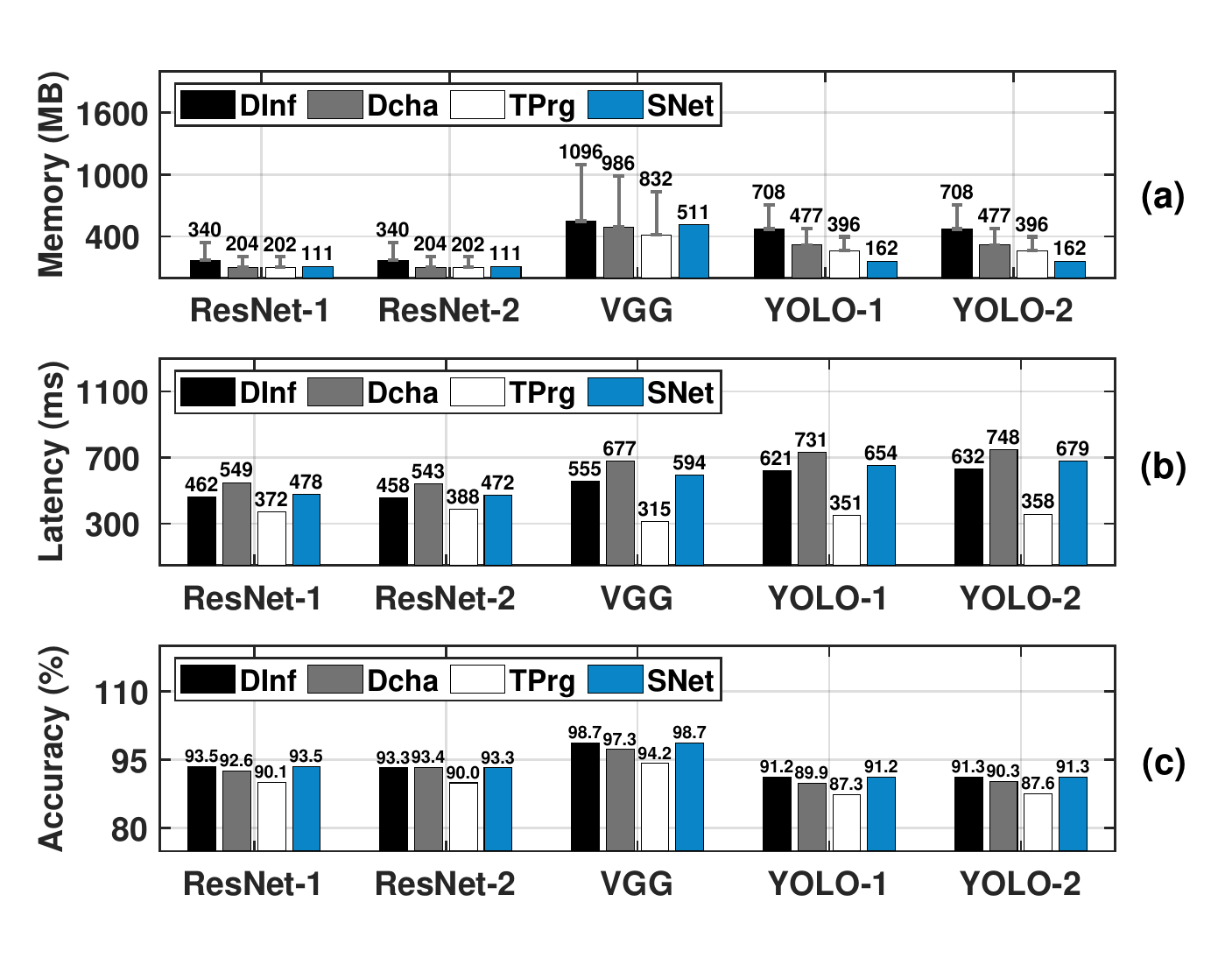}
	\caption{(a) Memory, (b) latency and (c) accuracy of each model in the road-side unit (RSU) application.}
	\label{f:overall2}
\end{figure}

\figref{f:overall1}(a) compares the memory consumption of each model using the three methods. 
VGG and ResNet are executed on CPU. 
When they are compressed by \compression, their model size with \compression is smaller than the other two methods. 
However, \compression and \sufficient use the page cache, doubling the peak memory consumption of each model, which is tolerated by the memory headroom. 
For YOLO and FCN (executed on GPU), these two methods also require one more CPU-to-GPU memory copying, which is retained by the framework during execution, tripling peak memory consumption. 
During the experiments, for evaluation purposes, we terminate some non-DNN tasks if there is insufficient memory to ensure that \sufficient and \compression can work properly. 
\systemname can avoid these memory overheads, and each model's memory consumption stays within its memory budget. 
Overall, SwapNet reduces the memory consumption by 56.9--82.8\%, 35.7--65.0\% and 42.0--66.4\% than DInf, TPrg and DCha, respectively.

In \figref{f:overall1}(b), due to the reduced model size, the latency of each model with \compression is smaller than the other two methods, but the accuracy drops by 5.0--6.7\%, as shown in \figref{f:overall1}(c). For DCha, it can keep the accuracy near the DInf and SNet but need more latency overhead due to it handles channels one by one and then combines them.
With efficient swapping and processing overhead reduction, the latency of \systemname is very close to that of \sufficient, such as only 26--46 ms slower. Because \systemname does not require model compression, \figref{f:overall1}(c) shows that each model with \systemname can maintain the same high accuracy as in \sufficient.

\fakeparagraph{Performance on Road-Side Unit}
In this application, five DNN models with a total size of 1360 MB need to execute within  memory budget of 1088 MB. 
The memory budgets according to our heuristic memory budget allocation algorithm is 119 MB for ResNet and 165 MB for YOLO, respectively. 
For the same reason as described in \figref{f:overall1}, the budget of VGG is increased to 520 MB. 
\figref{f:overall2}(a--c) compares the memory consumption, latency and accuracy of each task for the three methods. 
Similar to self-driving, SWapNet is also effective in reducing memory consumption in this application, outperforming by 53.4--77.1\%, 38.6--59.1\% and 45.6--66.0\% than DInf, TPrg and DCha respectively. Meanwhile, the latency of \systemname increases only 14--47 ms compared to \sufficient without compromising model accuracy.
\begin{figure}[t]
    \vspace{-.1in}
	\centering
	\includegraphics[width=3in]{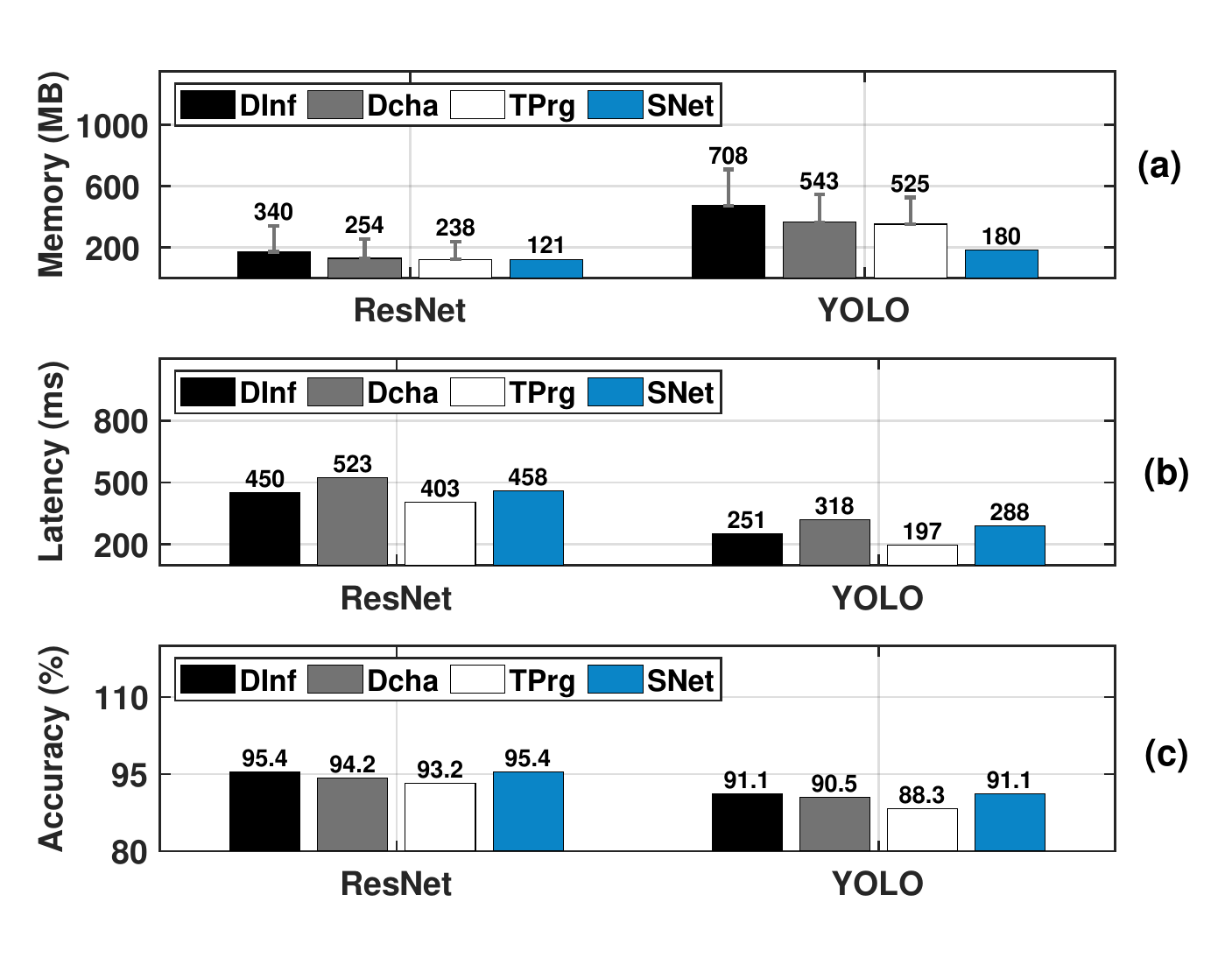}
	\caption{(a) Memory, (b) latency and (c) accuracy of each model in the UAV application.}
	\label{f:overall3}
\end{figure}

\fakeparagraph{Performance on UAV Surveillance}
In this application, we investigate a scenario with more memory resources, where the memory budget allocated to ResNet and YOLO is 136 and 189 MB memory budgets, respectively. 
In such a scenario, SwapNet can still effectively reduce memory overhead by 64.4--74.6\%, 49.2--65.7\% and 51.8--66.9\% compared to DInf, TPrg and DCha.
For latency, it is 8--37 ms slower than \sufficient without loss of accuracy in \figref{f:overall3}.
\looseness=-1
\begin{figure}[h]
	\centering
    \includegraphics[width=3in]{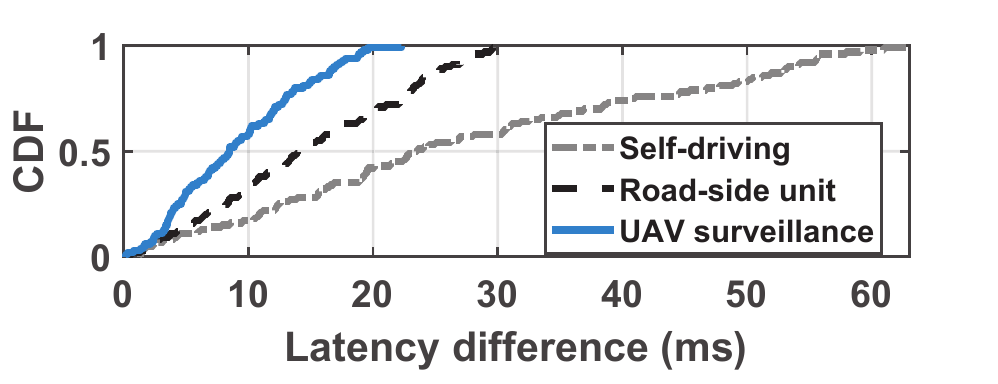}
	\caption{CDF of latency increases compared to \sufficient.}	
	\label{f:cdf}
\end{figure}

\fakeparagraph{Distribution of Latency Difference} 
To better understand the efficacy of the latency reduction by combining \systemname and the scheduling algorithm, we further investigate the distribution of the latency increases in \systemname compared to \sufficient. 
Due to page limitation, \figref{f:cdf} uses ResNet as an example to show the results, and the observations of other models are similar. 
In the application of self-driving, ResNet is divided into four blocks due to tight memory budget, while in RSU and UAV, it is divided into three blocks. 
For the same model, more blocks lead to greater latency because more blocks introduce more swapping and processing overheads, as shown in \figref{f:cdf} for  ``Self-driving''. 
On the other hand, even if a model is divided into the same number of blocks, the partition position may differ due to different memory budgets, which in turn leads to different latency. 
For example, the average latency difference of RSU is 5.5 ms smaller than that of UAV in \figref{f:cdf}.
\looseness=-1

\subsection{Ablation Study}
We develop three intermediate versions of \systemname to understand the efficacy for each of our proposed technical designs and discuss the potential benefits of combination the \systemname and proper scheduling algorithm.

\begin{itemize}
    \item \textbf{w/o-uni-add}: it removes the block swapping controller (\secref{sec:swap}) and falls back to memory copying.
    \item \textbf{w/o-mod-ske}: it removes block assembly controller (\secref{sec:assemble}), and uses existing model assembly in inference mode (\figref{f:restoration}(a)).
    
    \item \textbf{w/o-pat-sch}: it removes our simple scheduling scheme (\secref{sec:app:scheduling}) and uses a naive equal memory partition strategy.
\end{itemize}

\begin{figure}[t]
	\centering
	\includegraphics[width=3in]{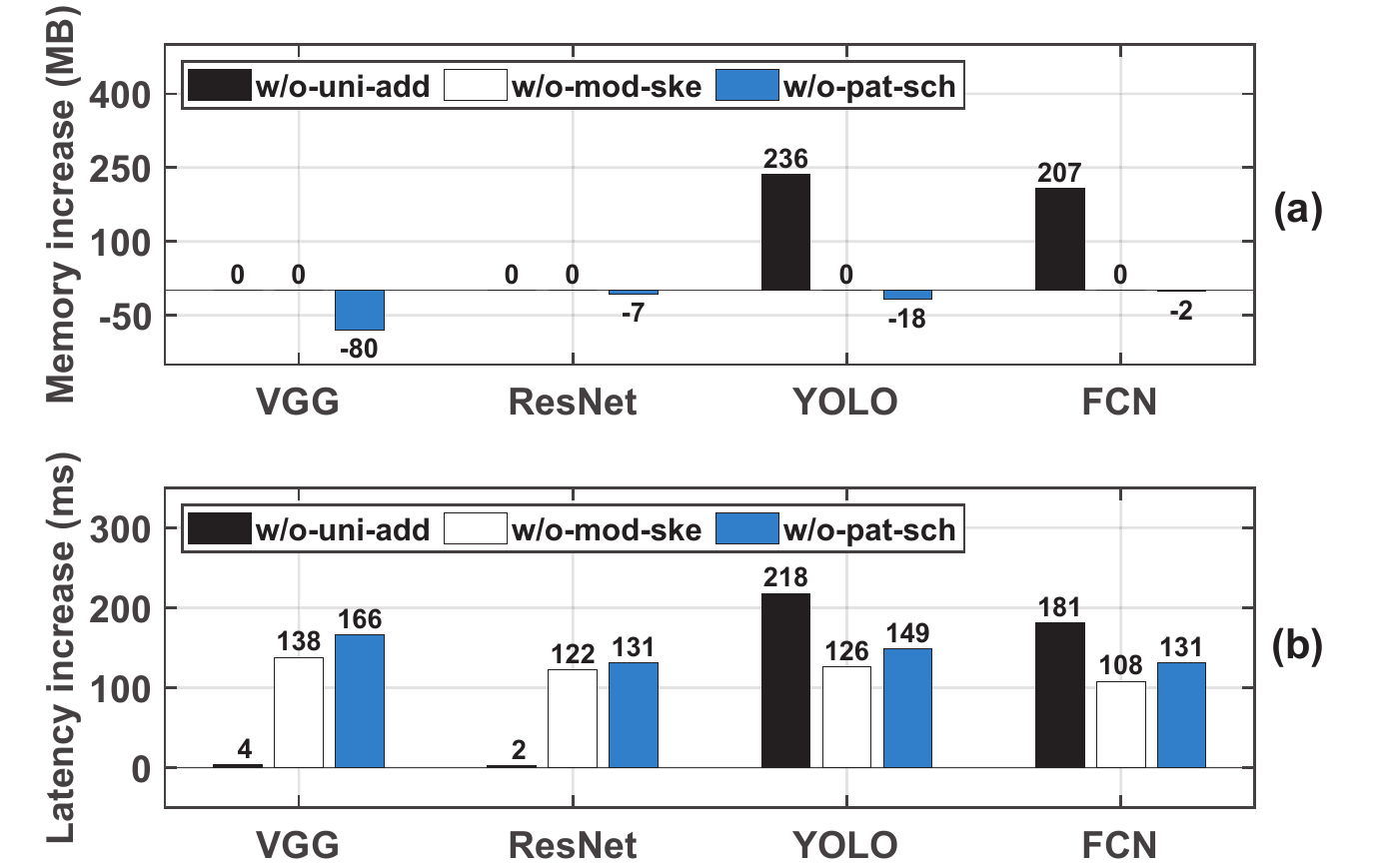}
	\caption{Ablation study of each intermediate system version for (a) memory  and (b) latency compared to that of the full version of \systemname.}
 \label{f:ablation1}
\end{figure}

From \figref{f:ablation1}, we can see that without unified addressing, ``w/o-uni-add'' greatly increases the memory consumption of models executed on GPU, such as YOLO and FCN. 
For clear representation, we plot the increased memory consumption of each intermediate version compared to that of the full version of \systemname, where negative values indicate memory reduction. 
The ``w/o-uni-add'' version also increases their latency by 26.3--50.1\% due to slow memory copying. 
Then, ``w/o-mod-ske'' increases the latency for all the models by 15.7--29.0\% when using the existing model assembly scheme. Since we use the assembly scheme in inference mode, ``w/o-mod-ske'' does not introduce additional memory overhead.
Finally, since our simple scheduling scheme focuses on reducing latency within the memory budget, there is a significant increase in latency by 19.0--34.3\% in ``w/o-pat-sch''. 
The results in \figref{f:ablation1} show the effectiveness of each of our technical designs in \systemname and also show the benefits from the combination of \systemname and suitable scheduling algorithm.

\vspace{-.1in}
\subsection{Micro-benchmarks}

This subsection investigates the performance of combining \systemname and our simple scheduling scheme under different system settings. 
We also discuss potential improvements for future scheduling algorithm designs.

\begin{figure}[ht]
	\centering
	\includegraphics[width=3in]{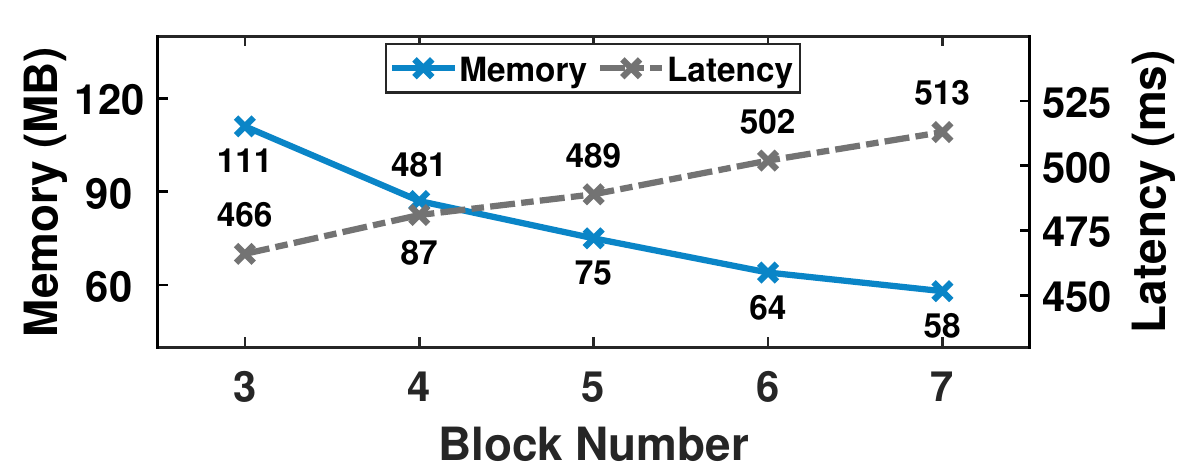}
	\caption{Memory and latency performance when the model is partitioned into more blocks.}
	\label{f:blcoknumber}
\end{figure}

\fakeparagraph{Impact of Block Numbers} 
In this experiment, we vary the number of model blocks. 
In \figref{f:blcoknumber}, ResNet is partitioned into 3 blocks by our simple scheduling scheme according to the memory budget, where the memory consumption is 111 MB and the execution latency of the model is 466 ms. 
We then intentionally partition the model with more block numbers, \eg, ranging from 4 to 7, so that each block tends to have a smaller block size. 

We can see that the memory consumption keeps decreasing as the number of blocks increases, since only two blocks co-exist in memory in the current \systemname design. 
The advantage is that this can further reduce memory requirements. 
Therefore, \systemname can be used as a solution to reduce the memory footprint of the application by controlling the memory requirements of the DNN model. 
However, the drawback is that latency will increase as each block introduces additional overhead. 
Therefore, it is recommended to keep the concurrency in 2 by default when using other potential scheduling algorithm, which can meet the memory budget and achieve low latency.

\begin{figure}[t]
	\centering
	\includegraphics[width=3in]{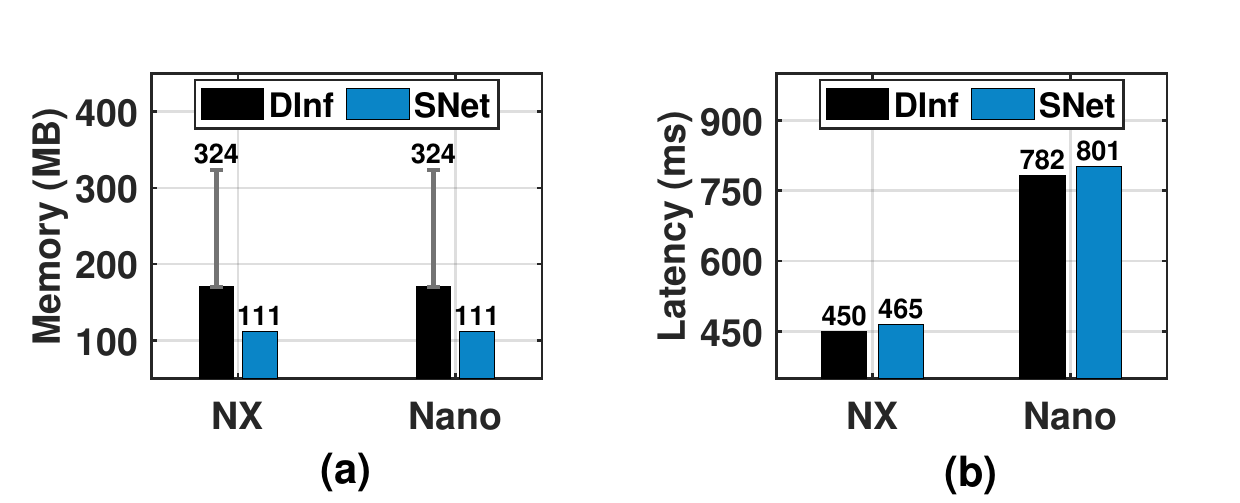}
	\caption{(a) Memory and (b) latency on different devices.}
	\label{f:device}
\end{figure}

\fakeparagraph{Performance on Different Devices} 
In this experiment, we deploy \systemname on a lower-end edge AI device Jetson Nano and examine the system performance. 
Given the same memory budget, we also run the same model (ResNet-101) on Jetson NX.
Hence, the scheduler provides the same partitioning, and their memory consumption is the same, \eg, 111 MB in \figref{f:device}(a). 
Although Nano has a slower CPU, the \systemname design is still effective. 
Compared to \sufficient, the latency is increased by 19 ms on Nano, similar to 15 ms on NX in \figref{f:device}(b).

\begin{figure}[h]
	\centering
	\includegraphics[width=2.8in]{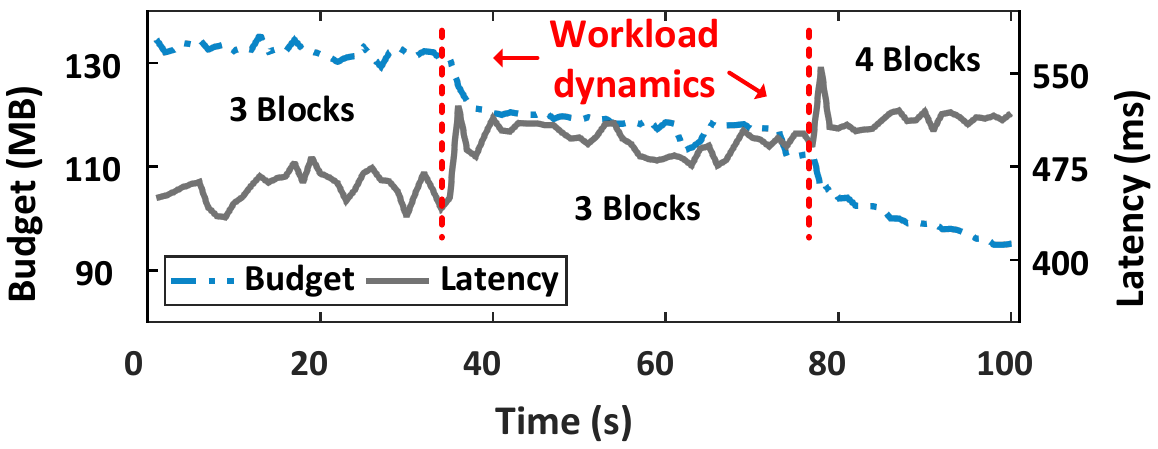}
	\caption{Runtime adaptation of model partitioning.}	
	\label{f:casestudy}
\end{figure}

\fakeparagraph{Adaptation to Dynamic Runtime Budget} 
In this experiment, we install \systemname on RosMaster X3 autonomous vehicle and apply a configuration similar to that of the self-driving application to process data collected from sensors in real time, as shown in \figref{f:scenario}. 
During device execution, we intentionally launch additional tasks to further reduce the available budget twice to investigate the robustness of \systemname to this dynamics. 
In \figref{f:casestudy}, we take ResNet (170 MB) as an example, starting with a memory budget of  136 MB, and the model is initially divided into three blocks.
Since the memory consumption of non-DNN tasks may vary slightly, the remaining available memory (used as a budget for DNN tasks) will vary accordingly.
\looseness=-1

Our simple scheduling scheme realize the fast adaptation according to compute several partition strategy look up tables before execution. During execution, it periodically reads the current memory budget. 
When the first workload dynamics occurs, the memory consumption of the initial model partition strategy exceeds the currently available budget, triggering the partition strategy adaptation. Adaptation is finished within 74 ms, where the model still has three blocks, but with new partition positions. 
The new strategy results in increased latency, \eg, about 499 ms on average. 
When the second workload dynamics occurs, the model is partitioned into four blocks and the adaptation can still be finished with a short delay, \eg, 64 ms. 
After this adaptation, the latency increases slightly to an average of 511 ms. 
This experiment shows that the combination of \systemname and our scheduling scheme can effectively respond to memory budget changes.
It also inspires other potential scheduling algorithms to integrate adaptive modules into the design.

\vspace{-.08in}
\subsection{System Overhead}

\begin{figure}[ht]
	\centering
	\includegraphics[width=3in]{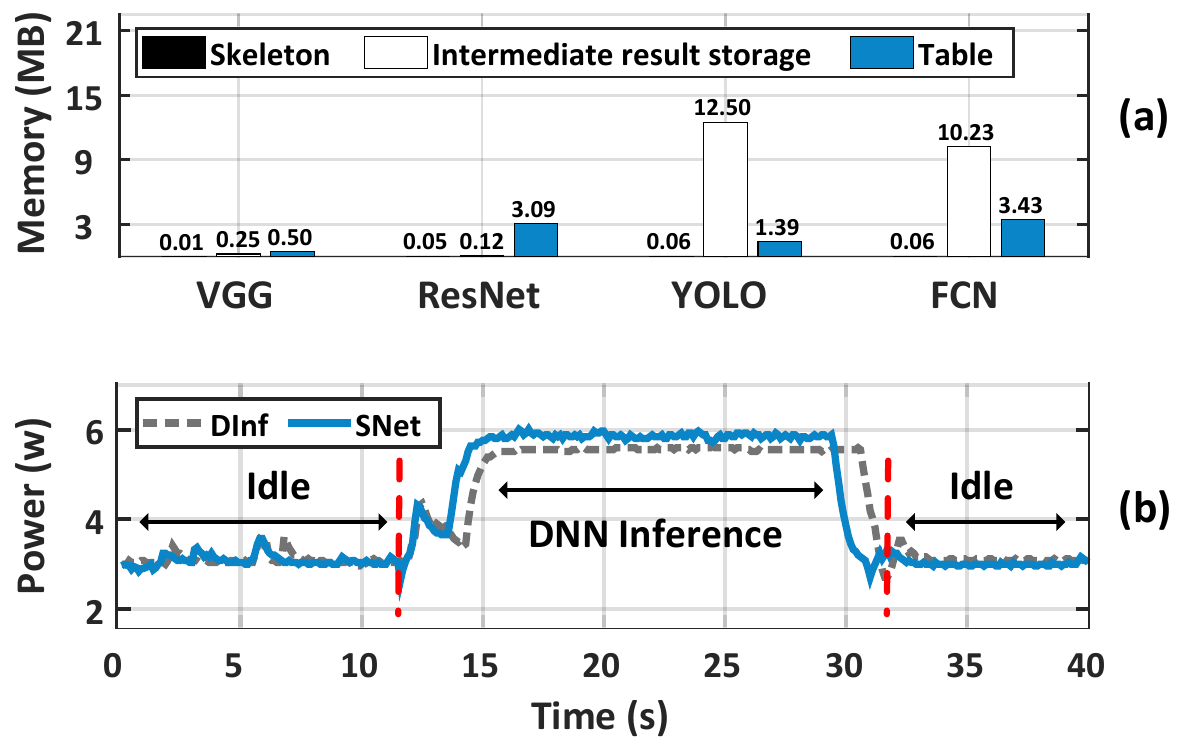}
	\caption{(a) Memory and (b) power overhead.}
	\label{f:overhead}
\end{figure}

\fakeparagraph{Memory Overhead} 
\systemname introduces three types of memory overhead, including model skeleton, temporal storage of feature values, and partition strategy tables, as shown in \figref{f:overhead}(a). 
In particular, each model used requires 0.01--0.06 MB of model skeleton and 0.12--12.50 MB of intermediate result storage. 
The strategy tables are 0.50--3.43 MB. Such memory overhead is 3.6\% on average, which is captured by the memory budget overhead $\delta$ in budget allocation. 

\fakeparagraph{Power Consumption} 
Low power consumption is important to edge AI devices, and we measure the power consumption of \systemname on Jetson NX using the INA3221 power monitor~\cite{power}. 
As a baseline, we also measure the power consumption of \sufficient without model partitioning. 
In \figref{f:overhead}(b), we can see that the device's idle state consumes about 3 W. 
When a model is running, the power consumption is 5.97 W (by \systemname) and 5.64 (by \sufficient), where \systemname itself consumes about 0.33 W only. 
Since our model assembly is faster, the blue curve is ahead of the grey curve.
\looseness=-1

\section{Related Work}
\label{sec:related}

\fakeparagraph{Model Execution with Memory Budget} 
To accommodate tight memory budgets, two popular solutions are explored in the literature: model compression \cite{han2015deep,liu2018demand,park2022mgemm,xie2019source,hou2022neulens,polino2018model,elsken2019neural,liu2021adaspring} and offloading \cite{kang2017neurosurgeon,jiang2022primask,huang2022real,lee2019occlumency,li2019edge,wang2021context}. 
Model compression techniques reduce the model size by removing redundant parameters such as layers, filters and channels \cite{han2015deep}, lowering the parameter precision \cite{polino2018model}, searching efficient model architectures \cite{elsken2019neural}, \etc 
However, when a model is compressed, its accuracy or robustness is often compromised, which is not favorable in mission-critical applications, \eg, self-driving.
Research on offloading dynamically changes model partition position \cite{wang2021context}, optimizes offloading patterns to reduce delay \cite{yao2020deep}, improves the inference privacy \cite{lee2019occlumency}, \etc 
Offloading does not harm model accuracy but requires network connections, making it vulnerable to network fluctuations. SwapNet do not have these disadvantages.

\fakeparagraph{Multi-DNN Execution} 
To efficiently support concurrent execution of multiple tasks, BlastNet~\cite{blastNet} achieves high resource utilization of heterogeneous CPU and GPU at runtime. 
RT-mDL~\cite{ling2021rt} supports real-time multi-DNN inference through joint model scaling and scheduling. 
Our \systemname is complementary by improving the execution efficiency if the memory demand is beyond the budget. 
It can be easily combined with these schemes to further improve the efficiency of multiple DNNs.

\fakeparagraph{CPU-GPU Co-processing on Mobile Devices} 
Recent studies mainly focus on improving the efficiency of DNNs on mobile devices by better scheduling of tasks and processors, such as the optimal task execution plan using AsyMo~\cite{wang2021asymo}, co-processing of CPU and GPU in CoDL~\cite{jia2022codl}, optimal graph-based task scheduling in Band~\cite{jeong2022band}, and memory layout optimization in Melon~\cite{wang2022melon}. 
Due to the difference in DNN development ecosystems for mobile and edge AI devices (see \secref{sec:pre:edge}), these solutions do not address the challenges encountered in \systemname.

\fakeparagraph{Existing Swapping Methods} 
A recent work~\cite{miao2021enabling} proposes to execute DNN models via block swapping. 
However, it is designed for MCUs, which has different design considerations and does not apply to edge AI devices.
SwapAdvisor~\cite{huang2020swapadvisor} also proposed the swapping strategy, but for traditional desktop-grade devices, which adopts a different memory architecture. These methods are not directly applicable to edge AI devices.

\section{Points of Discussion}

We present the following discussion related to this paper.

\textbf{1) Compared to other peer methods.} In the literature, model parallelism and partial computation offload~\cite{zhou2022accelerating,li2019edge} are also applied to cope with the memory shortages when running DNN models. Since they often require reliable network communication and support from other devices or servers, it may be problematic in the applications such as autonomous driving that often face unstable connections (\eg, crossing tunnels or switching base stations), raising safety concerns due to unpredictable response times. Moreover, they may also bring up data privacy concerns, as continuous data exchange with another device (or server) is needed. \systemname does not face these problems, but it is worth noting that the design of \systemname itself is actually orthogonal to these existing methods. Therefore, they can work together if the above two problems can be avoided in some applications, such as the device is always static with good network connection all the time and no sensitive data is involved.

\textbf{2) Potential future exploration.} The trend to handle complex language tasks in an increasing number of applications has promoted the deployment of large language models (LLMs) or their smaller approximations (such as LLaMA-7B)~\cite{llama} in the edge AI ecosystem. Therefore, this emerging trend requires the support and optimization of other model architectures, especially the transformers in LLMs~\cite{vaswani2017attention}, which is not considered in the current \systemname design. We will further investigate the compatibility of \systemname with operational flows and topologies of the transformer-like structures, which can provide novel and feasible insights for future deployment of LLMs on edge AI devices.

\vspace{-.1in}
\section{Conclusion}
\label{sec:concl}

This paper introduces \systemname, a middleware that logically executes large DNN models on a small memory budget. 
\systemname partitions large DNN models into blocks for execution by swapping them between the memory and the external storage in order.
Our main contribution is a transparent design that eliminates the substantial delay and memory overhead occurred during block swapping while remaining compatible with the DNN development tool chains for edge AI devices. Extensive evaluations show the promising performance gains of \systemname in combination with scheduling algorithms for efficient multi-DNN execution. 

\vspace{-.08in}
\section*{Acknowledgement}
This work is sponsored by the GRF grant from Research Grants Council of Hong Kong (Project No. CityU 11202623), the CityU SRG-Fd (Project No. 7005658) and APRC grant from City University of Hong Kong (Project No. 9610633). Zimu Zhou and Zhenjiang Li are the corresponding authors.

\let\oldbibliography\thebibliography
\renewcommand{\thebibliography}[1]{%
	\oldbibliography{#1}%
	\setlength{\itemsep}{1pt}%
}

\bibliographystyle{IEEEtran}
\bibliography{IEEEabrv,reference}
\vspace{-10mm}
\begin{IEEEbiography}
[{\includegraphics[width=1in,height=1.25in,clip,keepaspectratio]{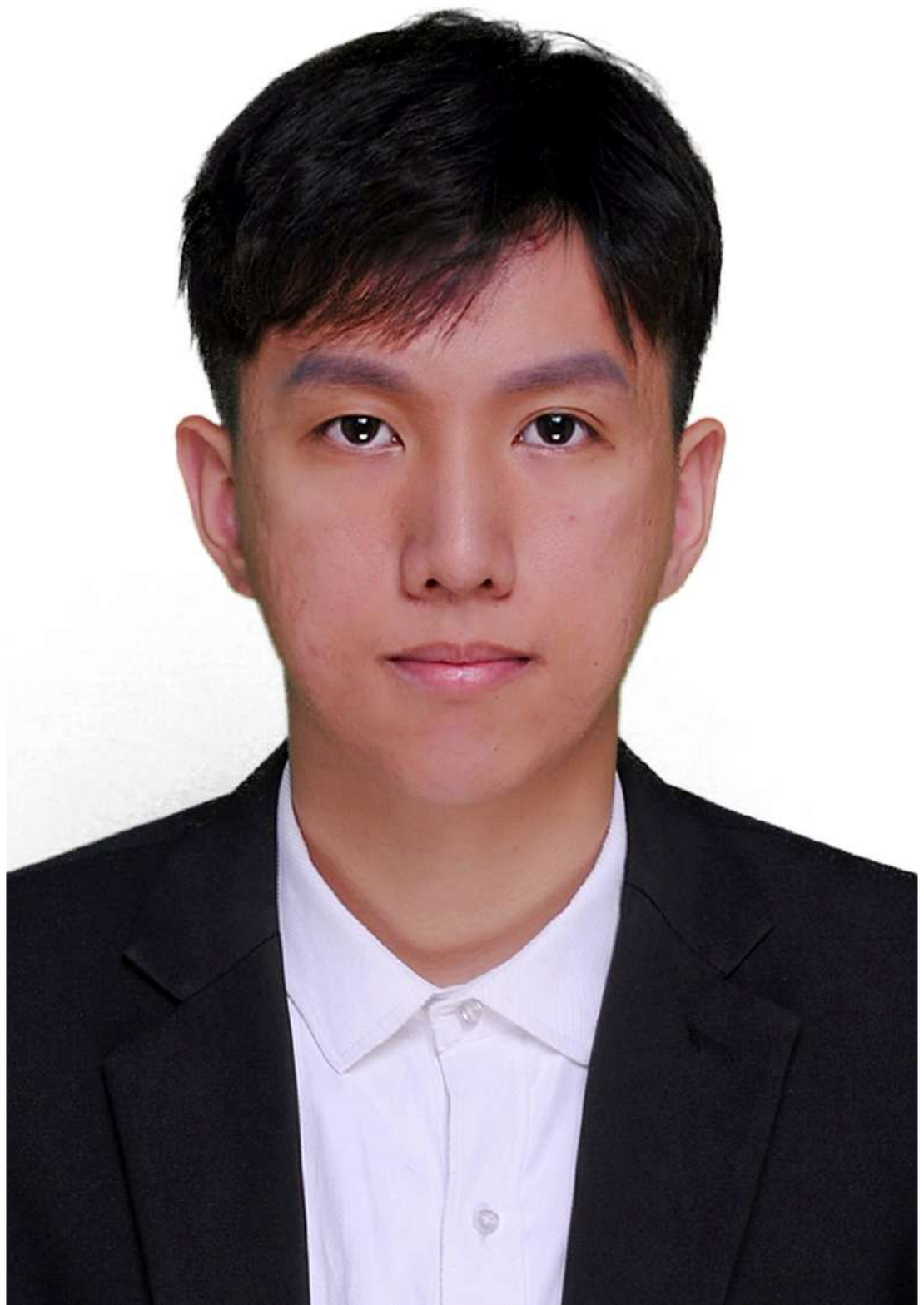}}]
{Kun Wang} received the B.E. degree from Xidian University, Xi’an, China, in 2020. He is currently working toward the PhD degree in the Department of Computer Science, City University of Hong Kong. His research interests include Edge AI system and Heterogeneous Federated Learning.\end{IEEEbiography}
\vskip -2.8\baselineskip plus -1fil
\vspace{-1mm}
\begin{IEEEbiography}
[{\includegraphics[width=1in,height=1.25in,clip,keepaspectratio]{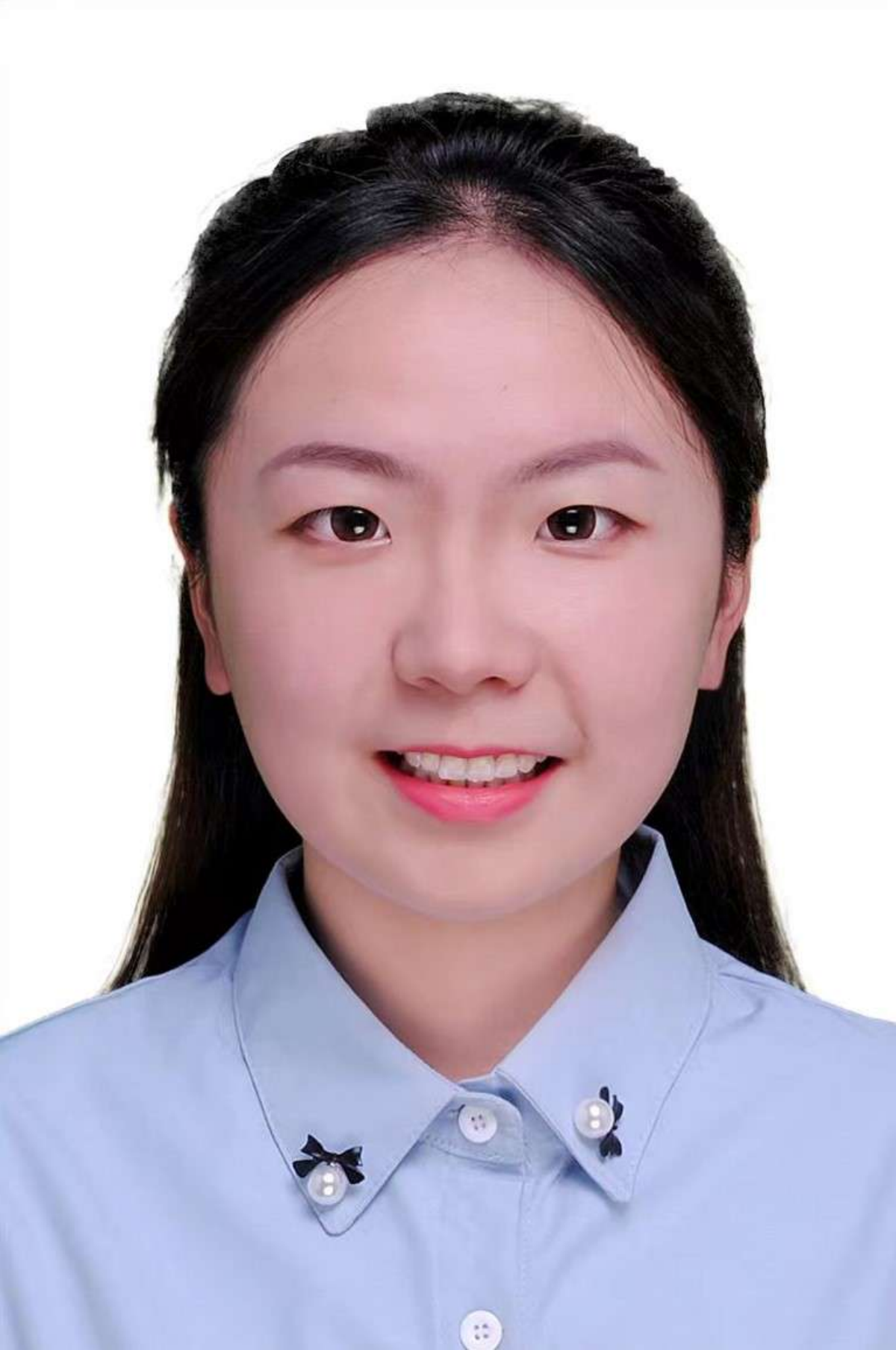}}]
{Jiani Cao} received the B.E. degree from Xidian University, Xi’an, China, in 2020. She is currently working toward the PhD degree in the Department of Computer Science, City University of Hong Kong. Her research interests include building Internet of Things (IoT) systems with signal processing and deep learning techniques to enable human-centric and VR/AR applications.\end{IEEEbiography}
\vskip -2.8\baselineskip plus -1fil
\begin{IEEEbiography}
[{\includegraphics[width=1in,height=1.25in,clip,keepaspectratio]{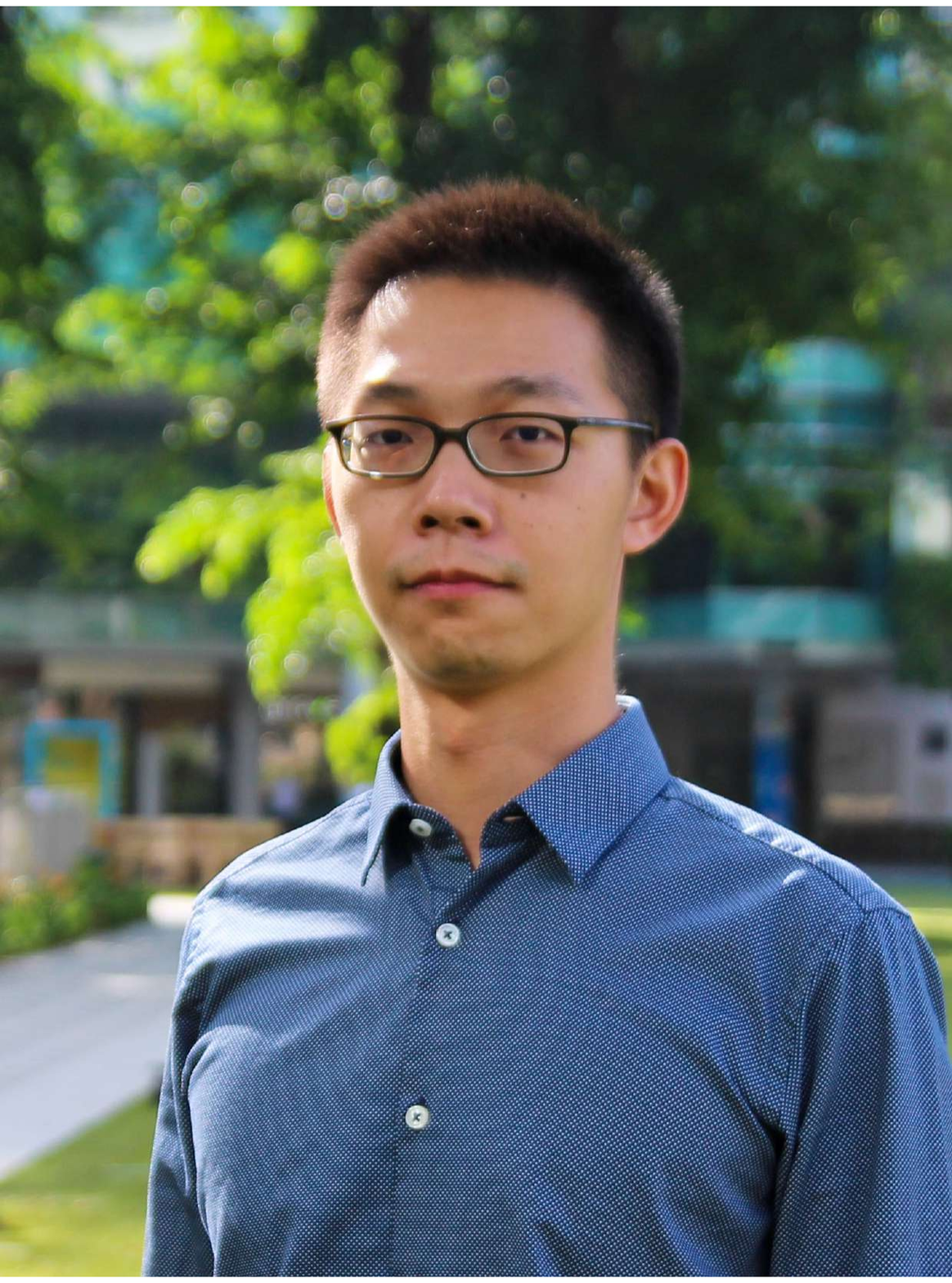}}]
{Zimu Zhou} received the B.E. from the Department of Electronic Engineering, Tsinghua University, Beijing, China, in 2011, and the Ph.D. from the Department of Computer Science and Engineering, Hong Kong University of Science and Technology, Hong Kong, in 2015. He is currently an Assistant Professor at the School of Data Science, City University of Hong Kong. His research focuses on mobile and ubiquitous computing.\end{IEEEbiography}
\vskip -2.8\baselineskip plus -1fil
\vspace{-2mm}
\begin{IEEEbiography}
[{\includegraphics[width=1in,height=1.25in,clip,keepaspectratio]{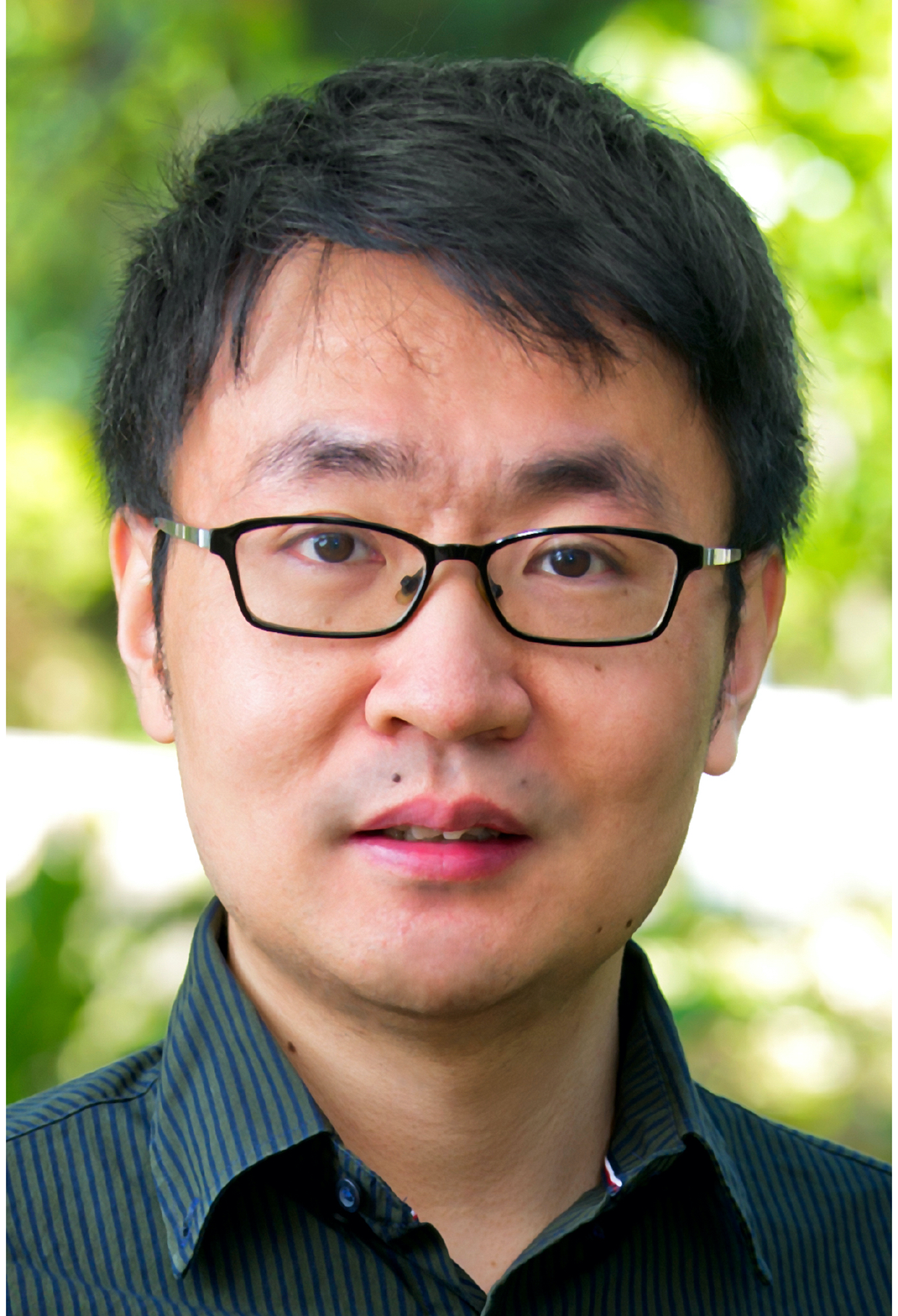}}]
{Zhenjiang Li} received the B.E. degree from Xi'an Jiaotong University, China, in 2007, and the M.Phil. and Ph.D. degrees from the Hong Kong University of Science and Technology, Hong Kong, China, in 2009 and 2012, respectively. He is currently an Associate Professor with the Department of Computer Science, City University of Hong Kong. His research interests include Internet of Things, edge AI systems and smart sensing.
\end{IEEEbiography}

\end{document}